\documentclass[10pt,twocolumn,letterpaper]{article}

\usepackage[pagenumbers]{iccv} 

\definecolor{iccvblue}{rgb}{0.21,0.49,0.74}
\usepackage[pagebackref,breaklinks,colorlinks,allcolors=iccvblue]{hyperref}

\usepackage{xcolor}         
\usepackage{enumitem}       
\usepackage{diagbox}        
\usepackage{array,multirow} 
\usepackage{algorithm}      
\usepackage{algorithmic}    
\usepackage{amsmath}        
\usepackage{amsfonts}       
\usepackage{graphicx}       
\usepackage{booktabs}       
\usepackage[export]{adjustbox}

\newcolumntype{L}[1]{>{\raggedright\let\newline\\\arraybackslash\hspace{0pt}}m{#1}} 
\newcolumntype{R}[1]{>{\raggedleft\let\newline\\\arraybackslash\hspace{0pt}}m{#1}}
\newcolumntype{C}[1]{>{\centering\let\newline\\\arraybackslash\hspace{0pt}}m{#1}}

\def\paperID{6987} 
\def\confName{ICCV}
\def\confYear{2025}

\title{Scheduling Weight Transitions for Quantization-Aware Training}

\author{
  Junghyup Lee$^{1,}$\thanks{Equal contribution.~$^\dagger$Corresponding author.} \quad\quad Jeimin Jeon$^{2,3,}$\footnotemark[1] \quad\quad Dohyung Kim$^{4}$ \quad\quad Bumsub Ham$^{2,\dagger}$\vspace*{0.2cm} \\
   $^{1}$ Samsung Research \hspace{8mm} $^{2}$Yonsei University \hspace{8mm} $^{3}$Articron Inc. \hspace{8mm} $^{4}$ Samsung AI center \\
   { \url{https://cvlab.yonsei.ac.kr/projects/TRS/}} 
}

\begin{document}
\maketitle

\begin{abstract}
    Quantization-aware training~(QAT) simulates a quantization process during training to lower bit-precision of weights/activations. It learns quantized weights indirectly by updating latent weights,~\ie,~full-precision inputs to a quantizer, using gradient-based optimizers. We claim that coupling a user-defined learning rate~(LR) with these optimizers is sub-optimal for QAT. Quantized weights transit discrete levels of a quantizer, only if corresponding latent weights pass transition points, where the quantizer changes discrete states. This suggests that the changes of quantized weights are affected by both the LR for latent weights and their distributions. It is thus difficult to control the degree of changes for quantized weights by scheduling the LR manually. We conjecture that the degree of parameter changes in QAT is related to the number of quantized weights transiting discrete levels. Based on this, we introduce a transition rate~(TR) scheduling technique that controls the number of transitions of quantized weights explicitly. Instead of scheduling a LR for latent weights, we schedule a target TR of quantized weights, and update the latent weights with a novel transition-adaptive LR~(TALR), enabling considering the degree of changes for the quantized weights during QAT. Experimental results demonstrate the effectiveness of our approach on standard benchmarks.
  \end{abstract}

  \section{Introduction}
Recent neural networks use wide and deep architectures~\cite{he2016deep,szegedy2016rethinking,huang2017densely,xie2017aggregated,hu2018squeeze} requiring millions of parameters and computations. Network quantization converts full-precision weights and/or activations into low-bit ones. This reduces storage usage and computational overheads drastically, but the low-bit models perform worse than the full-precision ones. To alleviate this problem, many approaches~\cite{rastegari2016xnor,choi2018pact,jung2019learning,esser2019learned,liu2020reactnet} adopt quantization-aware training~(QAT) that simulates a quantization process during training. It is not straightforward to optimize discrete quantized weights using gradient-based optimizers~(\eg, stochastic gradient descent~(SGD)) with continuous gradients. QAT instead exploits full-precision latent weights and a quantizer involving a discretization function~(\eg, a round function) to update the quantized weights indirectly.

QAT mainly consists of three steps: (1)~In a forward propagation step, full-precision latent weights are converted to quantized weights using a quantizer, and the quantized weights are used to compute an output; (2)~Gradients w.r.t an objective function are then back-propagated to the latent weights in a backward propagation step;~(3)~In an optimization step, the latent weights are updated with the gradients. Previous works mostly focus on the forward and backward propagation steps by designing quantizers~\cite{rastegari2016xnor,li2016ternary,zhou2016dorefa,cai2017deep,jung2019learning,esser2019learned,choi2018pact,zhang2018lq,yamamoto2021learnable} and addressing a vanishing gradient problem, caused by the discretization function in a quantizer~\cite{yang2019quantization,gong2019differentiable,kim2021distance,bai2018proxquant,lee2021network}, respectively. They do not pay attention to the optimization step, and simply exploit gradient-based optimizers with a user-defined learning rate~(LR), such as SGD or Adam~\cite{kingma2014adam}, to update the latent weights. This optimization strategy is, however, designed for training full-precision models, and does not consider how quantized weights are changed, which is sub-optimal for QAT.

\begin{figure*}[t]
    \captionsetup[subfigure]{justification=centering}
    \begin{center}
      \begin{subfigure}[t]{0.48\columnwidth}
        \centering
        \includegraphics[width=1\columnwidth]{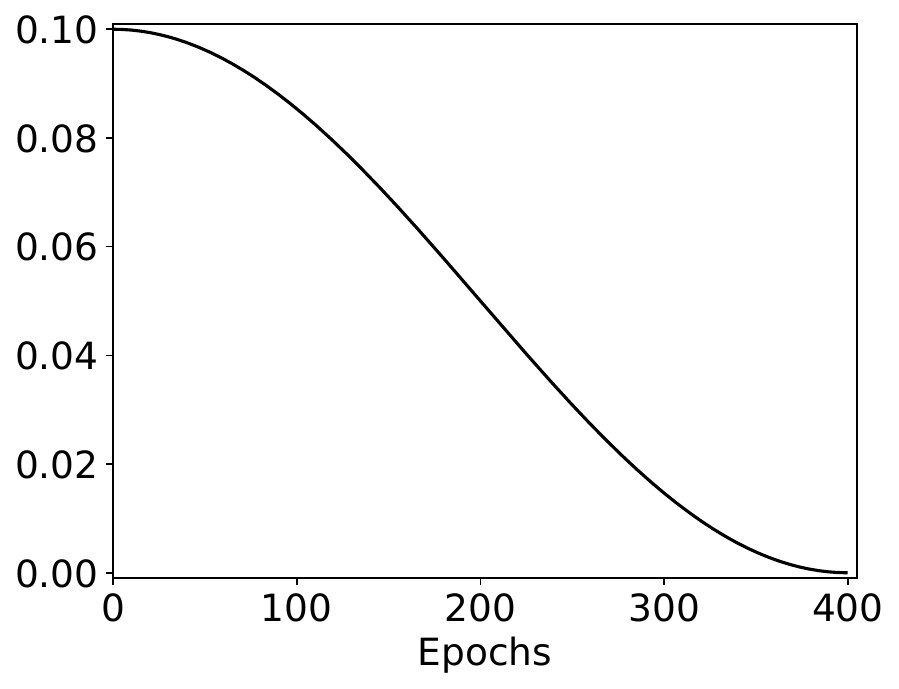}
        \caption{LR decay for SGD.}
        \label{fig:teaser_lr}
      \end{subfigure}
      \begin{subfigure}[t]{0.485\columnwidth}
        \centering
        \includegraphics[width=1\columnwidth]{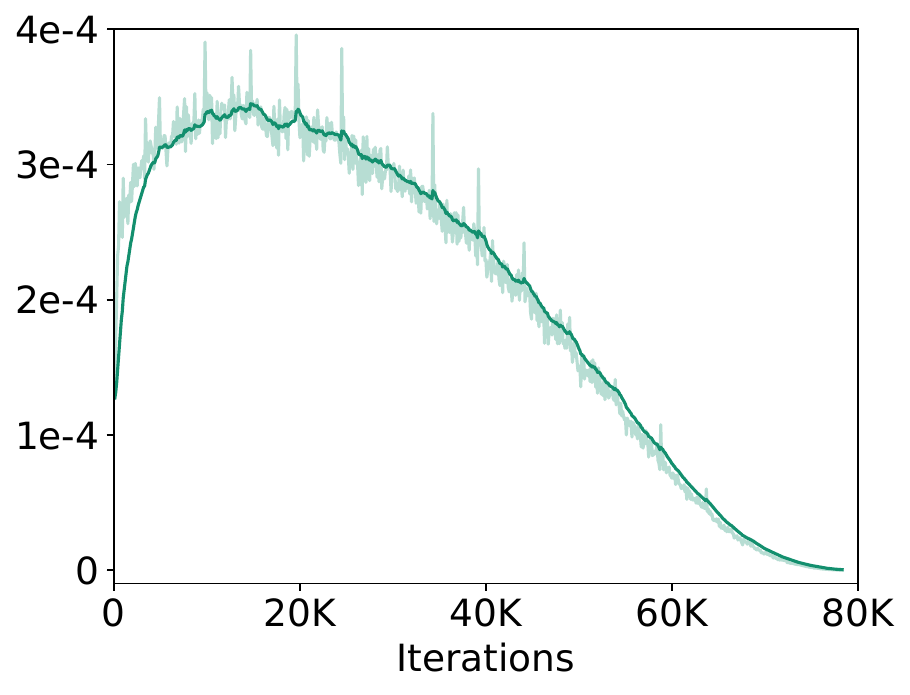}
        \caption{Avg. effective step sizes for FP weights (16$^\text{th}$ layer).}
        \label{fig:teaser_fp}
      \end{subfigure}
      \begin{subfigure}[t]{0.485\columnwidth}
        \centering
        \includegraphics[width=1\columnwidth]{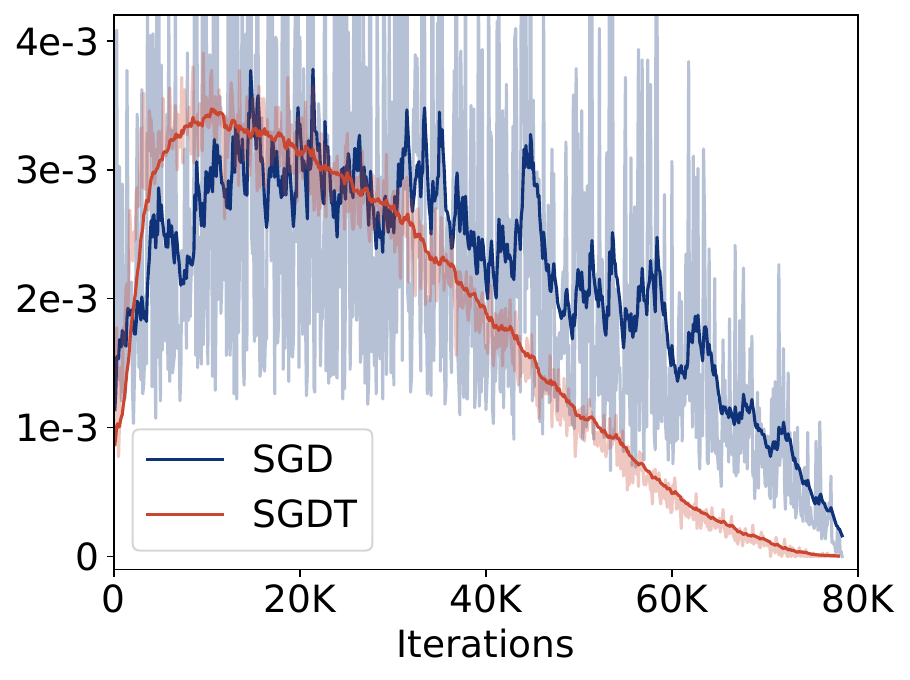}
        \caption{Avg. effective step sizes for 2-bit weights (16$^\text{th}$ layer).}
        \label{fig:teaser_quant}
      \end{subfigure}
      \begin{subfigure}[t]{0.48\columnwidth}
        \centering
        \includegraphics[width=1\columnwidth]{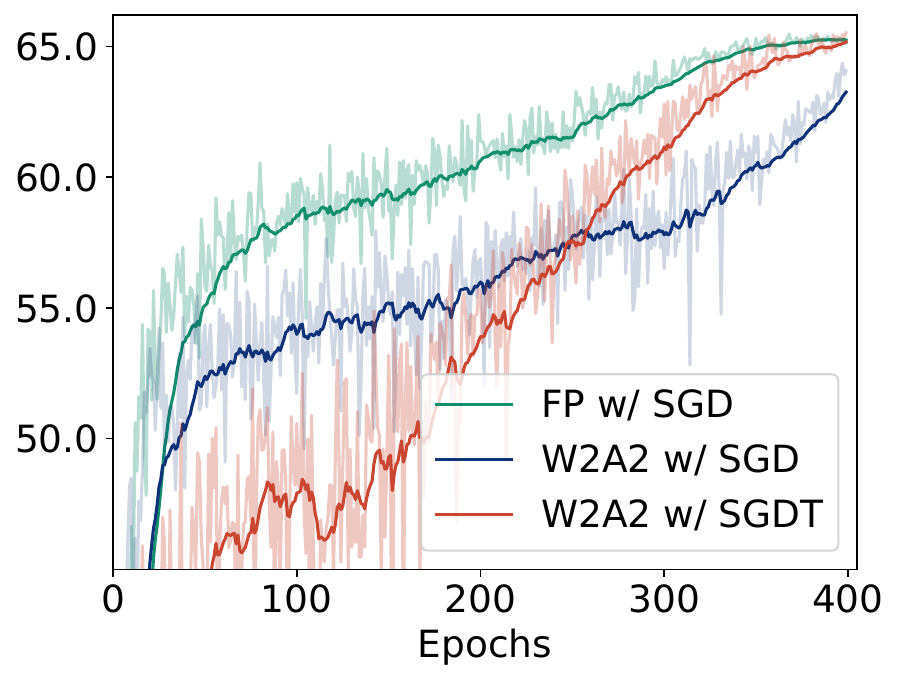}
        \caption{Test accuracy (\%).}
        \label{fig:teaser_acc}
      \end{subfigure}
    \end{center}
    \vspace{-0.65cm}
    \caption{Training curves of full-precision~(FP) and quantized models for ResNet-20~\cite{he2016deep} on CIFAR-100~\cite{krizhevsky2009learning}. Both weights~(W) and activations~(A) are quantized to a 2-bit precision~(W2A2). With a gradient-based optimizer~(SGD), we can control the average effective step size of FP weights roughly by scheduling a LR~({\subref{fig:teaser_lr} vs.~\subref{fig:teaser_fp}}), while we could not for quantized weights~(the blue curve in~{\subref{fig:teaser_quant}}). The curve for quantized weights is noisy, and decreases rapidly at the end of training, suggesting that 1) the quantized weights can alter significantly with a small LR and/or a small change of a LR, disturbing a coarse-to-fine parameter update and causing an unstable training, and 2) adopting a manually scheduled LR for QAT is sub-optimal. The optimizer coupled with our scheduling technique~(SGDT) can control the average effective step size of quantized weights by adjusting the number of transitions explicitly~(the red curve in~{\subref{fig:teaser_quant}}), showing better results in terms of accuracy and convergence~(the red curve in~{\subref{fig:teaser_acc}}).} \label{fig:teaser}
    \vspace{-0.6cm}
  \end{figure*}

When training a full-precision model with a gradient-based optimizer, we typically decay a LR progressively to update full-precision weights in a coarse-to-fine manner, which guarantees the convergence of the model~\cite{kleinberg2018alternative,huang2017snapshot}. That is, we can control the degree of weight changes manually by scheduling the LR, since the magnitude of a single parameter change, so-called an \emph{effective step size}~\cite{kingma2014adam}, is highly correlated with the LR~(Fig.~\ref{fig:teaser_lr} vs. Fig.~\ref{fig:teaser_fp}). For example, an average effective step size, which quantifies the degree of changes in weights, for a small LR is lower than that for a large one. We have found that this does not hold, when the optimizers coupled with a LR are applied to a quantized model in QAT. Since QAT updates quantized weights indirectly from latent weights and a quantizer, an average effective step size for quantized weights is less correlated with a LR for latent weights (Fig.~\ref{fig:teaser_lr} vs. the blue curve in Fig.~\ref{fig:teaser_quant}). To be more specific, different from full-precision weights, quantized weights change their discrete levels, only when corresponding latent weights pass transition points, where the quantizer alters discrete states~(we call this as \emph{transitions}). This makes it hard to control an average effective step size of quantized weights by adjusting a LR, disturbing coarse-to-fine parameter updates and the convergence of a model, even at the end of training~(the blue curve in Fig.~\ref{fig:teaser_acc}). For example, if latent weights are concentrated around a transition point, they can pass the point easily with a tiny LR, inducing significant changes of quantized weights.

  In this paper, we introduce a transition rate (TR) scheduling technique for QAT, which allows to update latent weights w.r.t the transitions of quantized weights explicitly. We define the TR of quantized weights as the number of quantized weights changing discrete levels at each iteration for optimization, divided by the total number of the weights. Note that an effective step size of each quantized weight is either zero or a discrete value~(\ie, a distance between two quantization levels), indicating that an average effective step size for quantized weights is mainly affected by the number of transitions. We thus conjecture that the number of transitions, or similarly, the TR is a key for controlling the degree of parameter changes for quantized weights. Based upon this, we propose to schedule a target TR of quantized weights, instead of a LR for latent weights, and update the latent weights with a novel transition-adaptive learning rate~(TALR) to adjust a TR of quantized weights accordingly. The TALR is changed adaptively to match the current TR of quantized weights with the target one. By scheduling the target TR, we are able to adjust the average effective step size of quantized weights~(the red curve in Fig.~\ref{fig:teaser_quant}). This allows us to optimize quantized weights in a coarse-to-fine manner, and provides a stable training process~(the red curve in Fig.~\ref{fig:teaser_acc}). To the best of our knowledge, scheduling a TR for QAT has not been explored, instead of a LR as in full-precision training. We demonstrate the superiority of our TR scheduling technique over the plain LR scheduling using various network architectures~\cite{sandler2018mobilenetv2,liu2020reactnet,he2016deep,touvron2021training,lin2017focal} and optimizers~(\eg, SGD, Adam~\cite{kingma2014adam}, and AdamW~\cite{loshchilov2017decoupled}) on image classification~\cite{krizhevsky2009learning,deng2009imagenet} and object detection~\cite{lin2014microsoft}. In summary, the main contributions of our work are threefold:

  \begin{itemize}[leftmargin=*]
    \item[$\bullet$] We claim the necessity of a training scheduler specialized for general QAT for the first time, where quantized weights are optimized indirectly by latent weights and a quantizer. To update latent weights in QAT, we propose to focus on the changes of quantized weights used for computing an output of a quantized network, which has not been covered by plain optimizers using a user-defined LR.
    
    \item[$\bullet$] We present a novel TR scheduling technique for QAT together with a TALR, which is adjusted considering a TR of quantized weights, controlling an average effective step size of quantized weights accordingly.
    
    \item[$\bullet$] We demonstrate the effectiveness and generalization ability of our approach on network quantization, boosting the performance of various models consistently.
  \end{itemize}
  

  \section{Related Work}

\paragraph{QAT.}
QAT methods simulate quantization during training by converting full-precision latent weights into quantized ones. Early works adopt fixed quantizers for binary~\cite{rastegari2016xnor,zhou2016dorefa}, ternary~\cite{li2016ternary}, and multi-bit~\cite{zhou2016dorefa,cai2017deep} representations, attempting to minimize quantization errors. Recent approaches introduce trainable quantizers that learn quantization parameters, such as intervals~\cite{jung2019learning,esser2019learned,choi2018pact} or non-uniform levels~\cite{zhang2018lq,yamamoto2021learnable}, significantly improving performance. These methods rely on the straight-through estimator~(STE)~\cite{bengio2013estimating} to handle non-differentiable quantization, but suffer from gradient mismatch~\cite{yang2019quantization,kim2021distance}. To address this, several works avoid exploiting the STE by using differentiable quantizers~\cite{yang2019quantization,gong2019differentiable,kim2021distance} or adjusting the gradients depending on the latent weights~\cite{bai2018proxquant,lee2021network}. Despite advancements, they overlook the fact that quantized weights act differently from full-precision counterparts, and simply adopt the same optimization strategies for full-precision models, that exploit manual LR scheduling techniques~(\eg, step decay or cosine annealing~\cite{loshchilov2016sgdr}), to train a quantized model. We highlight that quantized weights are updated indirectly by full-precision latent weights and a quantizer. Based on this, we propose to focus on actual changes of quantized weights, and present a TR scheduling technique specialized for QAT.

Recently, the work of~\cite{nagel2022overcoming} points out that quantized weights tend to oscillate between adjacent quantization levels during QAT. To address the oscillation problem, it proposes to freeze latent weights during training or to incorporate a regularization term into the objective function. While this approach could alleviate the oscillations, it still relies on the conventional optimization method using a LR, and thus it cannot control the average effective step size of quantized weights explicitly. Moreover, hyperparameters are chosen carefully according to network architectures and bit-widths, since freezing weights or adding the regularization term could disturb the training process. We have observed that it is difficult to control the average effective step size of quantized weights with the LR scheduling, and the reason for this is closely related to the oscillation problem. Different from~\cite{nagel2022overcoming} that reduces oscillations themselves by freezing or regularizing the weights, we attempt to control the actual change of quantized weights by scheduling a target TR. In this way, we can achieve consistent performance gains under the various bit-width settings with the same set of hyperparameters for each architecture, which is not feasible for~\cite{nagel2022overcoming}. 


Most recently, pseudo-quantization training (PQT)~\cite{defossez2021differentiable,savarese2022not,shin2023nipq} addresses oscillations by injecting pseudo-quantization noise into full-precision weights, instead of exploiting the quantization operations during training. While PQT stabilizes training, it introduces discrepancies between training and inference, as the actual quantization operations are not considered during training, leading to suboptimal performance. To overcome this, they leverage auxiliary techniques like knowledge distillation or mixed-precision quantization, but at the cost of computational overheads. In contrast, our method mitigates oscillations directly through TR scheduling, maintaining consistency between training and inference with minimal overhead, leading to consistent performance gains over plain optimization methods, with negligible overheads.

\vspace{-0.5cm}
\paragraph{Optimization Methods.}
Neural networks are generally trained using a gradient-based optimizer coupled with a LR scheduling technique. Gradient-based optimizers update network weights based on the gradients w.r.t an objective function. SGD is a vanilla optimizer for minimizing the objective function, which often exploits the first moment of the gradients to accelerate update steps near local optima. Many works~\cite{duchi2011adaptive,Tieleman2012,kingma2014adam} propose advanced optimizers using adaptive gradients. They accumulate squared gradients using a historical sum~\cite{duchi2011adaptive} or a running average~\cite{Tieleman2012,kingma2014adam}~(\ie, the second moment of gradients), which are then used for normalizing each gradient dimension adaptively. This enables highlighting infrequent features at training time, which is particularly useful for sparse gradients~\cite{duchi2011adaptive}. All of these optimizers exploit a LR to update full-precision weights. They typically use a large LR initially and decay it to a small value by a LR scheduling technique, such as step decay or cosine annealing~\cite{loshchilov2016sgdr}. The large initial LR encourages weights to explore local optima in a loss space, and the small LR at the end of training prevents the weights from overshooting from a local optimum~\cite{kleinberg2018alternative,huang2017snapshot}. A recent work~\cite{li2019towards} also points out that the LR scheduling is important for both generalization and performance, since large and small LRs play complementary roles in memorizing different types of patterns. The aforementioned optimization strategies are, however, designed for training a full-precision model, directly updating full-precision weights used for inference. For QAT, quantized weights are used for forward/backward passes, but full-precision latent weights are instead updated to train the quantized weights. We propose to consider this unique property of QAT to optimize the latent weights, and introduce a novel TR scheduling technique, specially designed for network quantization.

Closely related to ours, few works~\cite{helwegen2019latent,suarez2021bop} present optimizers for binary networks that train binary weights directly, rather than using latent weights. To this end, they adopt the first moment of gradients~\cite{helwegen2019latent} or its variant using the second moment~\cite{suarez2021bop} to flip binary values via thresholding. Although these methods do not use latent weights, the first moment and its variant have a role similar to the latent weights in that both accumulate gradients, and they schedule a threshold manually for flipping binary values. This suggests that they do not consider the degree of parameter changes in the binary weights explicitly, causing the same problem as the conventional optimizers using a LR.  These methods are also applicable for binary networks only. On the contrary, our method adjusts a TR explicitly to control an average effective step size of quantized weights consequently, and it can be applied to general QAT, including both binary and multi-bit quantization schemes.
\begin{figure*}[t]
    \captionsetup[subfigure]{justification=centering}
    \begin{center}
       \begin{subfigure}[t]{0.4\columnwidth}
          \centering
          \includegraphics[width=1\columnwidth]{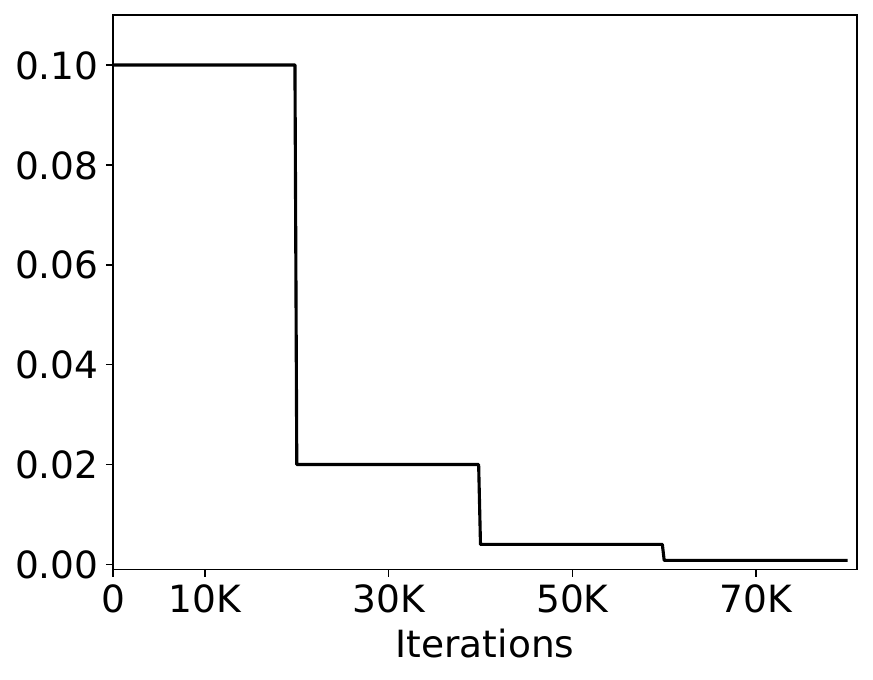}
          \caption{LR decay for SGD.}
          \label{fig:empirical_lr}
       \end{subfigure}
       \begin{subfigure}[t]{0.46\columnwidth}
          \centering
          \includegraphics[width=1\columnwidth]{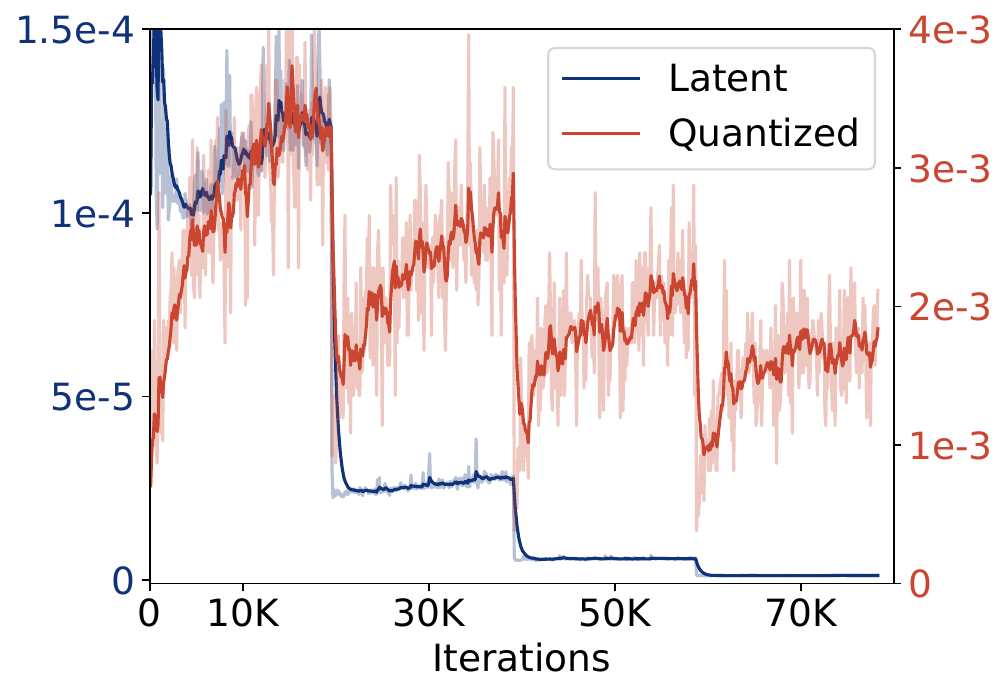}
          \caption{Avg. effective step sizes.}
          \label{fig:empirical_update}
       \end{subfigure}
       \begin{subfigure}[t]{0.48\columnwidth}
          \centering
          \raisebox{0.23cm}{\includegraphics[width=1\columnwidth]{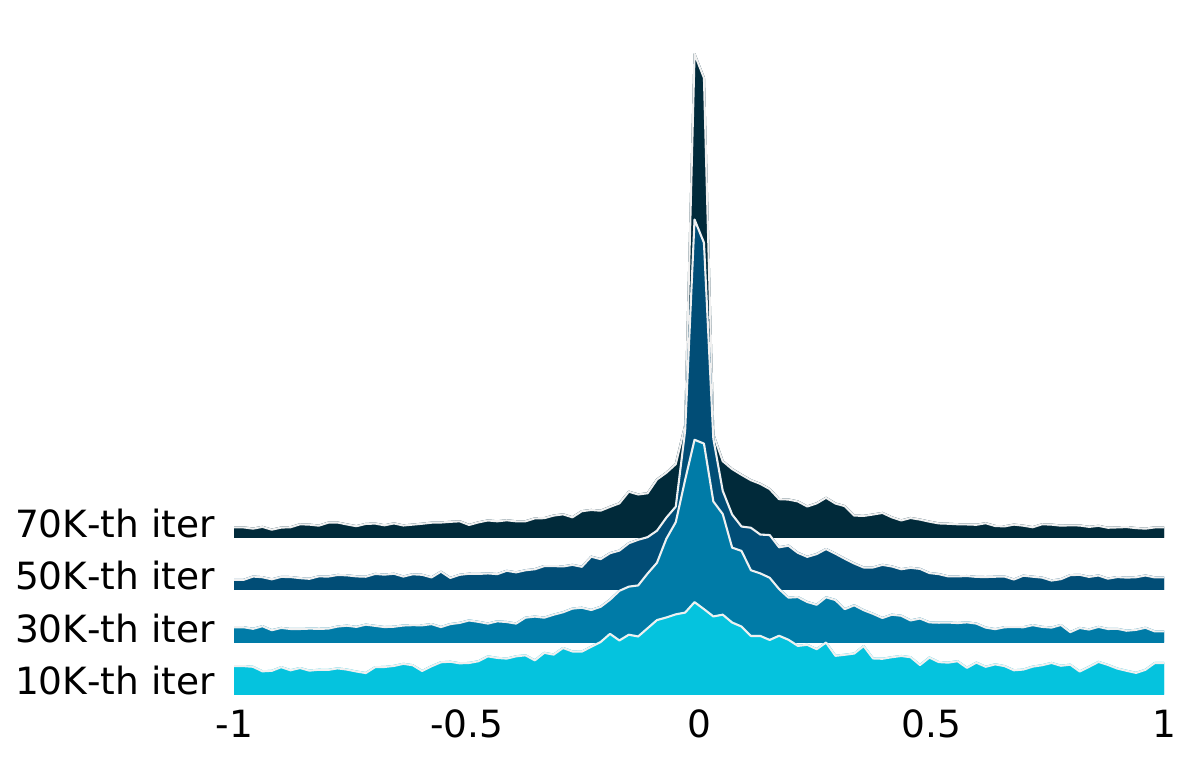}}
          \caption{Distribution of normalized \\ latent weights.}
          \label{fig:empirical_distribution}
       \end{subfigure}
       \begin{subfigure}[t]{0.40\columnwidth}
          \centering
          \includegraphics[width=1\columnwidth]{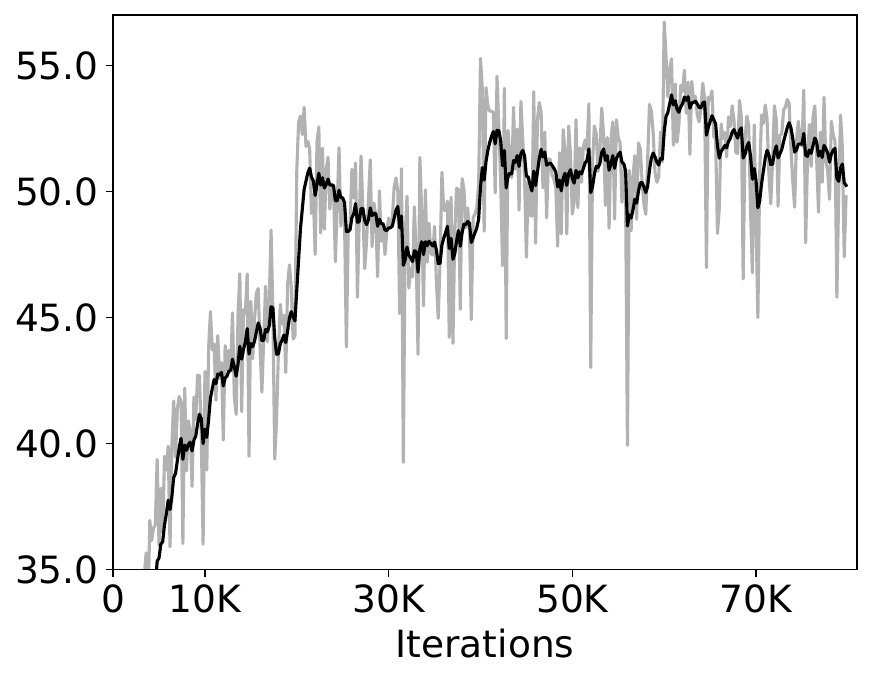}
          \caption{Test accuracy (\%).}
          \label{fig:empirical_acc}
       \end{subfigure}
    \end{center}
          \vspace{-0.6cm}
          \caption{Empirical analysis on QAT using SGD with a step LR decay. We binarize both weights and activations of ResNet-20~\cite{he2016deep} and train the model on CIFAR-100~\cite{krizhevsky2009learning}. For the visualizations in~{\subref{fig:empirical_update} and~\subref{fig:empirical_distribution}}, we track the latent and quantized weights in the 16$^\text{th}$ layer. We can see that the average effective step size of latent weights~(the blue curve in~{\subref{fig:empirical_update}}) is controlled by the LR in~{\subref{fig:empirical_lr}}, while that for the quantized weights changes significantly even with a small LR~(the red curve in~{\subref{fig:empirical_update}}). This is because the change of quantized weights is also affected by the distribution of latent weights approaching the transition point (\ie, zero in~{\subref{fig:empirical_distribution}}). The large changes in the quantized weights at the end of training~(the red curve in~{\subref{fig:empirical_update}}) degrade the performance in~{\subref{fig:empirical_acc}}. (Best viewed in color.)} 
          \vspace{-0.45cm}
    \label{fig:empirical}
  \end{figure*}

\section{Preliminary}
QAT inserts weight and/or activation quantizers into each layer of a neural network to simulate a quantization process at training time. Here we briefly describe a quantizer and an optimizer in QAT.

\vspace{-5mm}
\paragraph{Quantizer.}
Weight and activation quantizers take full-precision latent weights and activations in a layer, respectively, and produce low-bit representations. Here we mainly explain the weight quantizer. The activation quantizer is similarly defined. Let us denote by~${\bf{w}}$ full-precision latent weights. The quantizer first normalizes and clips the latent weights to adjust their range:
\vspace{-0.2cm}
\begin{equation} \label{eq:normalization}
   {\bf{w}}_n = f({\bf{w}}),
    \vspace{-0.1cm}
\end{equation}
where we denote by~${\bf{w}}_n$ normalized weights. $f$ is a normalization function involving scaling and clipping operations, which can be either hand-designed~\cite{zhou2016dorefa} or be trainable~\cite{jung2019learning,esser2019learned,choi2018pact}. The normalized weights~${\bf{w}}_n$ are then converted to discrete ones~${\bf{w}}_d$ using a discretization function~$g$:
\vspace{-0.2cm}
\begin{equation} \label{eq:discretization}
   {\bf{w}}_d = g({\bf{w}}_n).
   \vspace{-0.1cm}
\end{equation}
The discretization function~$g$ is typically a signum or a round function for binary or multi-bit quantization schemes, respectively. Note that STE~\cite{bengio2013estimating} is usually adopted in a backward pass to avoid a vanishing gradient problem, caused by the discretization function, propagating the same gradient from~${\bf{w}}_d$ to~${\bf{w}}_n$. Lastly, the quantizer produces quantized weights~${\bf{w}}_q$ by de-normalizing the discrete weights~${\bf{w}}_d$:
\vspace{-0.2cm}
\begin{equation} \label{eq:denormalization}
   {\bf{w}}_q = h({\bf{w}}_d),
    \vspace{-0.1cm}
\end{equation}
where~$h$ is a de-normalization function for post-scaling. The de-normalization could possibly be omitted (or fixed) when the quantized layer is followed by a normalization layer~(\eg, batch normalization~\cite{ioffe2015batch}), since it imposes the scale invariance to the weights and activations~\cite{hoffer2018norm,heo2021adamp}, suggesting that de-normalization has no effect on either the forward or backward pass.

\vspace{-5mm}
\paragraph{Optimizer.}
In QAT, the latent weights~${\bf{w}}$ are updated, instead of optimizing the quantized weights~${\bf{w}}_q$ directly. That is, updating the latent weights in turn alters the quantized ones during training. More specifically, the quantized weights change their discrete levels if corresponding normalized latent weights~${\bf{w}}_n$ pass transition points of the discretization function~$g$~(\eg, zero for the signum function) after updating the latent weights~${\bf{w}}$. Previous works typically use gradient-based optimizers with a user-defined LR to update the latent weights as follows:
\vspace{-0.2cm}
\begin{equation} \label{eq:conventional}
   {\bf{w}}^{t+1} = {\bf{w}}^{t} - \mu^{t} {\bf{g}}^{t},
    \vspace{-0.1cm}
\end{equation}
where the superscript~$t$ indicates an iteration step, and we denote by~${\bf{g}}$ and~$\mu$ a gradient term and the LR, respectively. Note that the gradient term~${\bf{g}}$ is computed differently depending on the types of optimizers. For example, SGD uses the first moment of gradients.


\vspace{-0.1cm}
\section{Method}
\vspace{-0.1cm}
In this section, we first present a detailed analysis of a conventional optimization method using a manually scheduled LR in the context of QAT~(Sec.~\ref{sec:problem}). We then introduce a novel TR scheduling technique~(Sec.~\ref{sec:TR_scheduler}).

\subsection{Empirical analysis} \label{sec:problem}
Conventional optimizers use a LR decay technique when training a full-precision model. They update model parameters gradually in a coarse-to-fine manner, which encourages a model to find a better local optimum in a loss space, and prevents overshooting from a local optimum~\cite{kleinberg2018alternative,huang2017snapshot}. This suggests that the optimizers control an average effective step size~(\ie,~the degree of parameter changes) of full-precision weights by adjusting the LR. We have empirically found that this does not hold for QAT. Namely, the average effective step size of quantized weights in QAT is hardly controlled by a conventional LR scheduling technique in gradient-based optimizers.

To understand this problem in detail, we show an empirical analysis on 1)~how a gradient-based optimizer, coupled with a manually scheduled LR, changes latent and quantized weights within a framework of QAT, and 2)~the influence of the changes on the classification accuracy of a quantized model~(Fig.~\ref{fig:empirical}). We train ResNet-20~\cite{he2016deep} with binary weights and activations on CIFAR-100~\cite{krizhevsky2009learning} using a SGD optimizer with a step LR decay method. We can see from Fig.~\ref{fig:empirical_lr} and the blue curve in Fig.~\ref{fig:empirical_update} that the average effective step size of latent weights is controlled by a LR, which is consistent with the result in a full-precision model~(\eg, Fig.~\ref{fig:teaser_lr} vs. Fig.~\ref{fig:teaser_fp}). The reason is that the latent weights in QAT and the weights in a full-precision model are continuous values, and the LR is responsible directly for updating the weights, \eg, as in Eq.~\eqref{eq:conventional}. On the contrary, quantized weights alter significantly, even with a small LR~(the red curve in Fig.~\ref{fig:empirical_update}). Since QAT uses quantized weights in a forward propagation step to compute gradients w.r.t an objective function, the large changes of quantized weights at the end of training make a training process unstable, disturbing a quantized model to converge~(Fig.~\ref{fig:empirical_acc}).

To delve deeper into this problem, let us suppose that a quantized weight needs to alter its discrete level~(\eg, from a negative value to a positive one in the binary quantization) in order to minimize a training loss. A corresponding latent weight then keeps accumulating gradients to move towards a transition point, and once a transition occurs, the latent weight might stay near the transition point. We can observe in Fig.~\ref{fig:empirical_distribution} that the normalized latent weights~(\ie, ${\bf{w}}_n$ in Eq.~\eqref{eq:normalization}) are approaching the transition point~(\ie, zero in this case) progressively according to the number of iterations. The quantized weight is hence likely to oscillate between adjacent discrete levels with small LRs in later training iterations (see the high peak at the 70K-th iteration in Fig.~\ref{fig:empirical_distribution}). This coincides with the recent finding in~\cite{nagel2022overcoming,park2020profit} that the quantized weights tend to oscillate during QAT, making it difficult to stabilize the batch normalization statistics~\cite{ioffe2015batch}, and degrading the performance at test time. This analysis indicates that 1)~the average effective step size of quantized weights is largely affected by the distribution of latent weights, and 2)~the reason why the LR is not a major factor for controlling the average effective step size in QAT, contrary to an optimization process of a full-precision model, is that the quantized weight alters only when the latent weight passes a transition point of a quantizer, but the LR cannot adjust the number of transitions explicitly. Consequently, our empirical analysis suggests the necessity of a training scheduler specific to QAT that allows to update latent weights adaptively considering the transitions in quantized weights.

\subsection{TR scheduler} \label{sec:TR_scheduler}
Here we present a relationship between an effective step size and transitions in quantized weights, and describe our approach to TR scheduling in a single layer.

\vspace{-5mm}
\paragraph{TR of quantized weights.}
We say that a transition occurs if a latent weight passes a transition point of a quantizer after a single update. The number of transitions is hence equal to that of quantized weights changing discrete levels after the update. We can count the number of transitions by observing whether discrete weights~(\ie,~${\bf{w}}_d$ in Eq.~\eqref{eq:discretization}) are changed or not after the update. Here we focus on a TR, the number of transitions divided by the total number of quantized weights, defined as follows:
\begin{equation} \label{eq:TR}
  k^t = \frac{\sum^{N}_{i=1} \mathbb{I}\left[ w^t_d (i) \neq w^{t-1}_d (i) \right]}{N},
\end{equation}
where we denote by~$k^t$ and~$w^t_d (i)$ the TR and the~$i$-th element of discrete weights at the~$t$-th iteration step, respectively, and~$N$ is the total number of quantized weights. $\mathbb{I}[\cdot]$ is an indicator function that outputs one if a given statement is true and zero otherwise.

\vspace{-4mm}
\paragraph{Relation between an effective step size and a transition.}
An \emph{effective step size}~\cite{kingma2014adam} indicates the magnitude of a single parameter change. We can compute the effective step size of a quantized weight~$w_q$ by measuring its absolute difference before and after a single update as follows:
\begin{equation} \label{eq:ESS}
  \left\vert \triangle w^t_q \right\vert = \left\vert w^t_q - w^{t-1}_q \right\vert,
\end{equation}
where we denote by~$\vert \triangle w^t_q \vert$ an effective step size of the quantized weight at the~$t$-th iteration step. We will show that the effective step size is related to a transition of the quantized weight. Let us denote by~$\delta^t$ a post-scaling factor of the de-normalization function $h$ in Eq.~\eqref{eq:denormalization}  at the~$t$-th iteration step. If the discretization function~$g$ in Eq.~\eqref{eq:discretization} is a rounding function for multi-bit quantization~(\ie~a discrete weight $w_d^t$ is an integer value), we can rewrite Eq.~\eqref{eq:ESS} as follows:
\begin{equation}
  \left\vert \triangle w^t_q \right\vert = \left\vert \delta^t w^t_d - \delta^{t-1} w^{t-1}_d \right\vert.
\end{equation}
If~$g$ is a signum function~(\ie,~$w_d^t \in \{-1, 1\}$) for binary quantization, Eq.~\eqref{eq:ESS} can be represented as follows:
\begin{equation}
  \left\vert \triangle w^t_q \right\vert = \frac{1}{2} \left\vert \delta^t w^t_d - \delta^{t-1} w^{t-1}_d \right\vert.
\end{equation}
Note that the change of $\delta^t$ in a single update is typically small~(\ie,~$\delta^t \approx \delta^{t-1}$) or we can set the post-scaling factor~$\delta^t$ as a constant value if the quantized layer is followed by a normalization layer~\cite{hoffer2018norm,heo2021adamp}~(\eg,~as in~\cite{lee2021network}). Assuming that the change of~$\delta^t$ is negligible within a single update and a latent weight passes a single transition point when a transition occurs, we can approximate the effective step size of the quantized weight as follows:
\begin{equation} \label{eq:stepsize}
  \left\vert \triangle w^t_q \right\vert \approx \delta^t \mathbb{I}\left[ w^t_d \neq w^{t-1}_d \right].
\end{equation} 
That is, the effective step size of the quantized weight is at most~$\delta^t$ if a transition occurs, and zero otherwise. This indicates that individual effective step sizes of quantized weights are discrete values~(\ie, zero or~$\delta^t$) determined by the quantizer. Note that the effective step size for each full-precision weight can be adjusted by a LR, since the weight is a continuous value, which is however not applicable for the quantized weight changing discretely. Accordingly, adjusting the number of transitions, or equivalently a TR, is important to control an average effective step size of quantized weights. Based upon this, we design a TR scheduling technique adjusting a TR of quantized weights explicitly, allowing us to control the degree of parameter changes in the quantized weights accordingly.

\vspace{-4mm}
\paragraph{TR scheduler.}
We incorporate our TR scheduling technique into an optimization process by introducing a transition-adaptive learning rate~(TALR) to update latent weights, allowing to adjust a TR of quantized weights manually, w.r.t a target TR. To this end, we mainly apply three operations at every iteration: Estimating a running TR using a momentum estimator, adjusting a TALR w.r.t a target value, and updating latent weights. Specifically, we first compute a running TR of quantized weights for each iteration~$t$ using an exponential moving average with a momentum of~$m$:
\begin{equation} \label{eq:EMA_tr}
   K^t = m K^{t-1} + (1-m)k^t,
\end{equation}
where we denote by~$K^t$ a running TR. Motivated by the running statistics in \eg, batch normalization~\cite{ioffe2015batch}, we use the momentum estimator to obtain the running TR, which roughly averages the TRs over recent training iterations, instead of using the TR,~$k^t$ in~Eq.~\eqref{eq:TR}, directly. This allows us to use a stable statistic of the TR, and alleviates the influence from outliers. We then adjust a TALR based on the running TR~$K^t$ and a target one:
\begin{equation} \label{eq:TALR}
   U^t = \max \left( 0, U^{t-1} + \eta \left( R^t - K^t \right) \right),
\end{equation}
where we denote by~$U^t$ and~$R^t$ the TALR and the target TR at the iteration step~$t$, and $\eta$ is a hyperparameter controlling the extent of the TALR update. Note that we can schedule the target TR~$R^t$ using typical schedulers~(\eg, step decay), which is analogous to the LR scheduling technique. With the TALR~$U^t$ at hand, we update the latent weights~${\bf{w}}^t$ as follows:
\begin{equation} \label{eq:update}
   {\bf{w}}^{t+1} = {\bf{w}}^t - U^t {\bf{g}}^{t},
\end{equation}
where ${\bf{g}}^{t}$ is a gradient term computed depending on the type of an optimizer~(\eg, the first moment of gradients in SGD). Updating the latent weights~${\bf{w}}^t$ with the TALR~$U^t$ enables controlling the running TR of quantized weights~$K^t$ w.r.t the target TR~$R^t$. For example, if a current running TR~$K^t$ is smaller than the target one~$R^t$, the TALR~$U^t$ increases according to Eq.~\eqref{eq:TALR}. The latent weights in Eq.~\eqref{eq:update} are then updated largely, compared to the previous iteration. This encourages more latent weights to pass transition points of a quantizer, which in turn raises the TR in the next step. Similarly, in the opposite case, the TALR decreases to reduce the TR. Note that one can adjust the TALR in a different way from Eq.~\eqref{eq:TALR} while achieving the same effect, and we discuss the variants of update algorithms for TALR in the Sec.~S3.2 of the supplement. Our approach connects the latent and quantized weights, in contrast to conventional optimization methods, making it possible to control an average effective step size of quantized weights via scheduling a target TR. 

\subsection{Quantization scheme}  \label{sec:scheme}
We apply the TR scheduler to QAT with various bit-width settings, including binary and multi-bit representations. In the following, we describe quantization schemes used in our experiments.

\vspace{-0.5cm}
\paragraph{Multi-bit quantization.}
We modify LSQ~\cite{esser2019learned}, the state-of-the-art method for multi-bit uniform quantization\footnote{Using the same network architecture~(\ie,~a vanilla version of ResNet), our modifications provide similar or better baseline results on ImageNet~\cite{deng2009imagenet}, compared to the performance of LSQ, reproduced in~\cite{bhalgat2020lsq+}.}. We define our $b$-bit quantizer as follows:
\begin{equation} \label{eq:multibit}
   {\bf{x}}_q = \frac{1}{\gamma} \left\lceil \text{clip} \left( \frac{\gamma {\bf{x}}}{s}, \alpha, \beta \right) \right\rfloor,
\end{equation}
where ${\bf{x}}_q$ is an output of the quantizer. We denote by~${\bf{x}}$ an input to the quantizer, which can be either latent weights or input activations. $\text{clip}(\cdot,\alpha,\beta)$ is a clipping function with lower and upper bounds of~$\alpha$ and~$\beta$, respectively, and $\lceil \cdot \rfloor$ is a round function. Following LSQ, we employ a learnable scale parameter~$s$ for each quantizer, adjusting the range of quantization interval\footnote{We train scale parameters in activation quantizers only, and do not train them in weight quantizers, when the TR scheduling technique is adopted. Otherwise, transitions could occur, even when the latent weights are not updated. For a fair comparison, we use learnable scale parameters for weight quantizers, when using plain optimizers without TR scheduling. See the Sec.~S5.2 of the supplement for details.}. We set the bit-specific constants~$(\alpha,~\beta,~\gamma)$ as $\left(  -2^{b-1},~2^{b-1}-1,~2^{b-1} \right)$ and $\left( 0,~2^{b}-1,~2^{b} \right)$ for weight and activation quantizers, respectively. We do not perform a post-scaling with the learnable scale parameter~$s$ after the round function in contrast to LSQ. That is, we fix the output range of a quantizer, enforcing the output of the quantizer~${\bf{x}}_q$ to be fixed-point numbers, regardless of the range of an input~${\bf{x}}$, which is more suitable for hardware implementation. Note that the scale difference between the input and output of a quantizer does not matter if each convolutional/fully-connected layer is followed by a normalization layer (\eg, batch normalization~\cite{ioffe2015batch}), imposing the scale invariance after every quantized layer~\cite{hoffer2018norm,heo2021adamp}. This ensures that post-scaling does not affect either the forward or backward pass. When the normalization is not used, we optionally apply a learnable post-scaling technique to outputs of convolutional/fully-connected layers~\cite{lee2021network}. 

\vspace{-0.4cm}
\paragraph{Binary quantization.}
We apply two binarization methods. First, we use the network architecture of ReActNet~\cite{liu2020reactnet} and its quantization scheme, which is the state of the art on binary quantization. ReActNet modifies the ResNet~\cite{he2016deep} or MobileNet-V1~\cite{howard2017mobilenets} architectures by adopting the Bi-Real structure~\cite{liu2018bi} that adds more residual connections, while exploiting real-valued $1 \times 1$ convolutions in the residual connections. This approach also uses learnable shift operations before quantization and activation functions. Second, we binarize vanilla ResNet models to compare binary and multi-bit quantization schemes under a fair training setting. To this end, we design a binary  quantizer using Eq.~\eqref{eq:multibit}. For a weight quantizer, we set $\alpha$, $\beta$, and $\gamma$, to -1, 1, and 1, respectively, and replace the round operator with a signum function to obtain a binary value of $-1$ or $1$. For an activation quantizer, we set those values as 0, 1, and 1, respectively, to generate a binary activation of $0$ or $1$.

\section{Experiments}
We describe our experimental settings~(Sec.~\ref{sec:settings}) and show results on image classification and object detection~(Sec.~\ref{sec:results}). We then analyze the TR scheduling technique~(Sec.~\ref{sec:analysis}). More detailed analyses and discussions are provided in the supplement.

\begin{table}[t]
   \setlength{\tabcolsep}{0.55em}
   \centering
   
   \caption{Quantitative comparison of quantized models on ImageNet~\cite{deng2009imagenet} in terms of a top-1 validation accuracy. We train quantized models with plain optimization methods~(SGD and Adam~\cite{kingma2014adam}) or ours using a TR scheduler~(SGDT and AdamT). The bit-widths of weights (W) and activations (A) are represented in the form of W/A. For comparison, we report the performance of full-precision~(FP) and activation-only binarized~(W32A1) models. The results of ReActNet-18~\cite{liu2020reactnet} for the plain optimizers are reproduced with an official source code.}
   \small
   \vspace{-0.3cm}
   \adjustbox{max width=0.49\textwidth}{
    \begin{tabular}{C{1.4cm} c c c  c c c c c}
       \midrule
       \multirow{3}{*}{Optimizer} & \multicolumn{3}{c}{MobileNetV2} & \multicolumn{1}{c}{ReActNet-18}   & \multicolumn{4}{c}{ResNet-18}  \\
                                        & \multicolumn{3}{c}{(FP: 71.9)}  & \multicolumn{1}{c}{(W32A1: 66.8)} & \multicolumn{4}{c}{(FP: 69.9)} \\
                                        & 2/2 & 3/3 & 4/4                 & 1/1                               & 1/1 & 2/2 & 3/3 & 4/4          \\
       \cmidrule(lr){1-1} \cmidrule(lr){2-4} \cmidrule(lr){5-5} \cmidrule(lr){6-9}
       SGD   & 46.9      & 65.6      & 69.9                   & 65.0                              & 55.3      & 66.8      & 69.5      & 70.5 \\
       SGDT  & \bf{53.6} & \bf{67.0} & \bf{70.5}              & \bf{65.3}                         & \bf{55.8} & \bf{66.9} & \bf{69.7} & \bf{70.6} \\
       \cmidrule(lr){1-1} \cmidrule(lr){2-4} \cmidrule(lr){5-5} \cmidrule(lr){6-9}
       Adam  & 49.6      & 66.5      & 70.0                   & 65.3                              & 56.1      & 66.7      & 69.5      & 70.1 \\
       AdamT & \bf{53.8} & \bf{67.3} & \bf{70.8}              & \bf{65.7}                         & \bf{56.3} & \bf{67.2} & \bf{69.7} & \bf{70.4} \\
       \midrule
     \end{tabular}} \label{tab:imagenet_main}
     \vspace{-0.3cm}
 \end{table}
 
 \begin{table}[t]
   \setlength{\tabcolsep}{0.55em}
   \centering
   \caption{Quantitative comparison of quantized models on CIFAR-100/10~\cite{krizhevsky2009learning} in terms of a top-1 test accuracy.}
    \small
   \vspace{-0.3cm}
   \adjustbox{max width=0.49\textwidth}{
    \begin{tabular}{c c c c  c c c}
       \midrule
       \multirow{4}{*}[-3pt]{Optimizer} &  \multicolumn{3}{c}{CIFAR-100}                                      & \multicolumn{3}{c}{CIFAR-10}  \\
       \cmidrule(lr){2-4} \cmidrule(lr){5-7}
       \multicolumn{1}{c}{}             & \multicolumn{1}{c}{ReActNet-18}   & \multicolumn{2}{c}{ResNet-20}  & \multicolumn{1}{c}{ReActNet-18}   & \multicolumn{2}{c}{ResNet-20}  \\
       \multicolumn{1}{c}{}             & \multicolumn{1}{c}{(W32A1: 69.6)} & \multicolumn{2}{c}{(FP: 65.1)} & \multicolumn{1}{c}{(W32A1: 91.3)} & \multicolumn{2}{c}{(FP: 91.1)} \\
                                        & 1/1                               & 1/1       & 2/2                & 1/1                               & 1/1       & 2/2 \\
       \cmidrule(lr){1-1} \cmidrule(lr){2-2} \cmidrule(lr){3-4} \cmidrule(lr){5-5} \cmidrule(lr){6-7}
       SGD                              & 69.7                              & 54.9      & 64.1               & 90.9                              & 85.2      & 90.2 \\
       SGDT                             & \bf{72.2}                         & \bf{55.8} & \bf{65.5}          & \bf{93.0}                         & \bf{85.6} & \bf{90.7} \\
       \cmidrule(lr){1-1} \cmidrule(lr){2-2} \cmidrule(lr){3-4} \cmidrule(lr){5-5} \cmidrule(lr){6-7}
       Adam                             & 69.5                              & 54.8      & 63.3               & 90.4                              & 84.8      & 90.2 \\
       AdamT                            & \bf{71.8}                         & \bf{55.9} & \bf{65.2}          & \bf{92.9}                         & \bf{85.7} & \bf{91.1} \\
       \midrule
     \end{tabular}} \label{tab:cifar_main}
     \vspace{-0.45cm}
 \end{table}

 \begin{table}[t]
   \centering
   \caption{Quantitative comparison of quantized models on ImageNet~\cite{deng2009imagenet} in terms of a top-1 validation accuracy. We train quantized models with plain optimization method~(AdamW~\cite{loshchilov2017decoupled}) or ours using a TR scheduler~(AdamWT).}
   \small
   \vspace{-0.3cm}
   \adjustbox{max width=0.95\columnwidth}{
     \begin{tabular}{C{1.6cm} C{1.4cm} C{1.4cm} C{1.4cm}  C{1.4cm} }
       \midrule
       \multirow{3}{*}{Optimizer} & \multicolumn{2}{c}{DeiT-T} & \multicolumn{2}{c}{DeiT-S}\\
                                        & \multicolumn{2}{c}{(FP: 72.0)}  & \multicolumn{2}{c}{(FP: 79.9)} \\
                                        & 2/2 & 3/3                & 2/2 & 3/3                                \\
       \cmidrule(lr){1-1} \cmidrule(lr){2-3} \cmidrule(lr){4-5}
       AdamW   & 54.6      & 68.1    & 68.4      & 77.6 \\
       AdamWT  & \bf{57.4} & \bf{69.5} & \bf{71.8}              & \bf{78.5}                          \\
       \midrule
     \end{tabular}} \label{tab:imagenet_vit}
     \vspace{-0.3cm}
 \end{table}

 \begin{table}[t]
   \setlength{\tabcolsep}{0.55em}
   \centering
   \small
   \caption{Quantitative results on object detection. We train RetinaNet~\cite{lin2017focal} on the training split of MS COCO~\cite{lin2014microsoft} using either the plain optimization method~(SGD) or ours~(SGDT). We report the average precision~(AP) on the validation split.}
   \vspace{-0.3cm}
   \adjustbox{max width=0.49\textwidth}{
    \begin{tabular}{c c c c c c c c c}
       \midrule
       Backbone                         & W/A                   & Optimizer & AP         & AP$_{50}$  & AP$_{75}$  & AP$_{S}$   & AP$_{M}$   & AP$_{L}$ \\
       \midrule 
       \multirow{5}{*}[-3pt]{ResNet-50} & FP                    & SGD       & 37.80      & 57.62      & 40.50      & 23.12      & 41.39      & 49.70 \\ \cmidrule{2-9}
                                        & \multirow{2}{*}{4/4}  & SGD       & 38.05      & 57.75      & 40.23      & \bf{22.51} & 41.46      & 49.68 \\
                                        &                       & SGDT      & \bf{38.36} & \bf{58.01} & \bf{40.76} & 22.46      & \bf{41.87} & \bf{49.71} \\ \cmidrule{2-9}
                                        & \multirow{2}{*}{3/3}  & SGD       & 37.32      & 56.87      & 39.71      & \bf{21.90} & 40.82      & 48.97 \\
                                        &                       & SGDT      & \bf{37.59} & \bf{56.89} & \bf{40.18} & 21.51      & \bf{40.98} & \bf{49.07} \\ 
 
       \midrule
     \end{tabular}} \label{tab:ms_coco_main}
     \vspace{-0.4cm}
 \end{table}

\begin{figure*}[t]
   \captionsetup[subfigure]{justification=centering}
   \begin{center}
      \begin{subfigure}[t]{0.45\columnwidth}
         \centering
         \includegraphics[width=1\columnwidth]{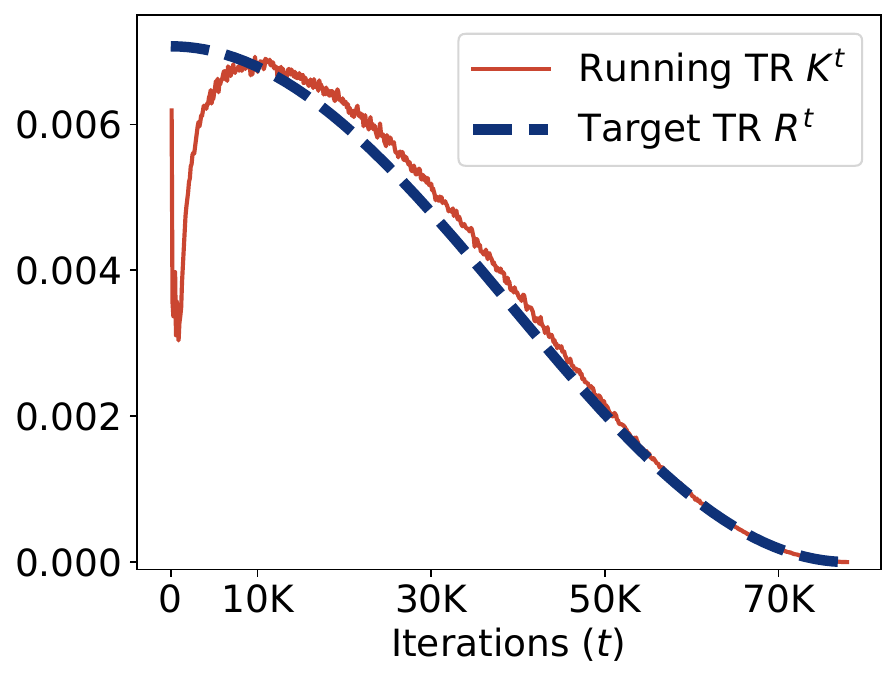}
         \caption{Running TR~$K^t$ and \newline target TR~$R^t$.}
         \label{fig:discussion_tr}
      \end{subfigure}
      \begin{subfigure}[t]{0.426\columnwidth}
         \centering
         \includegraphics[width=1\columnwidth]{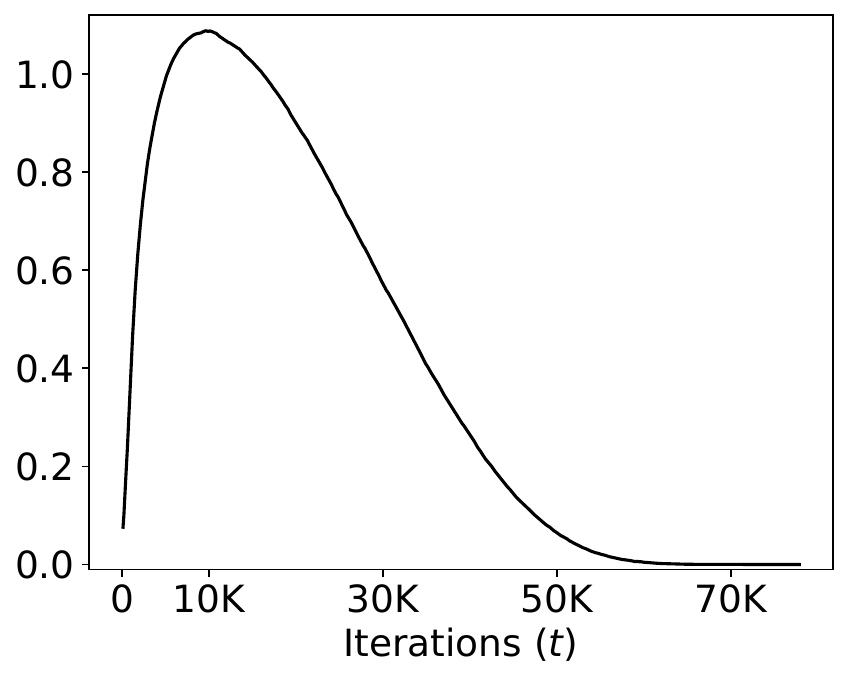}
         \caption{TALR~$U^t$.}
         \label{fig:discussion_talr}
      \end{subfigure}
      \begin{subfigure}[t]{0.495\columnwidth}
         \centering
         \raisebox{0.148cm}{\includegraphics[width=1\columnwidth]{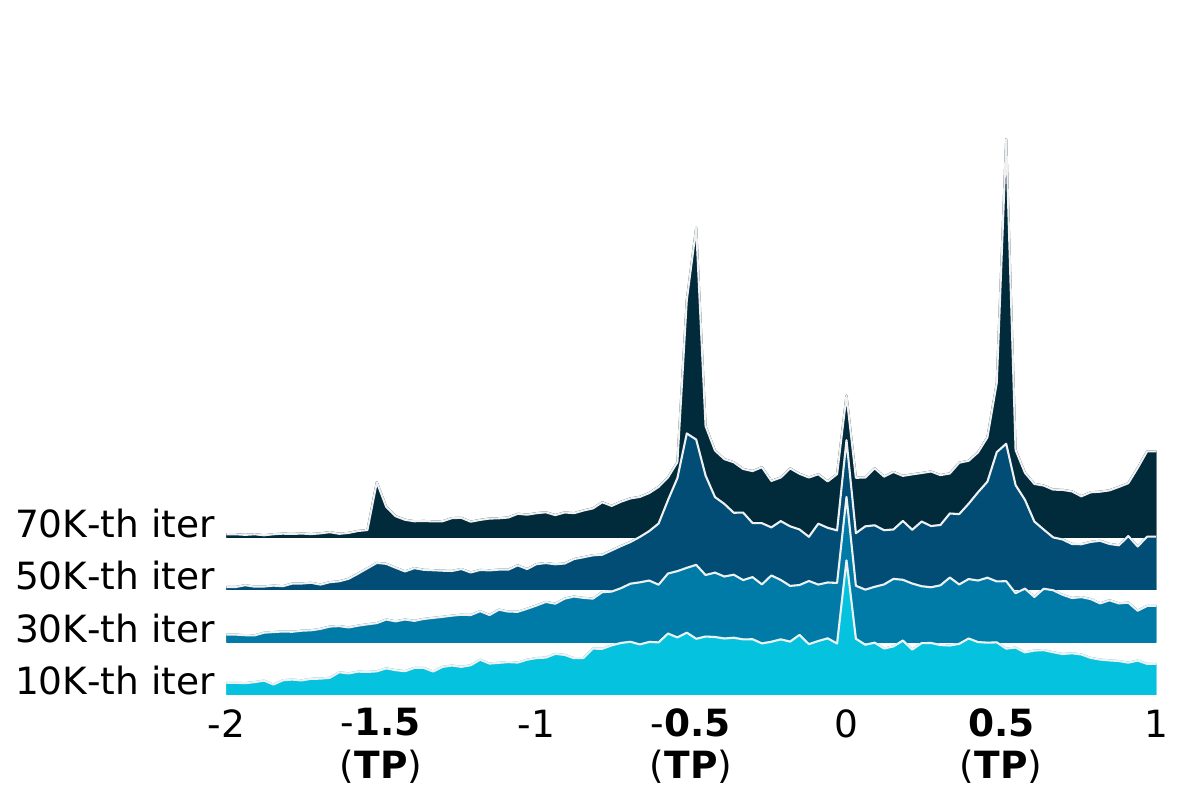}}
         \caption{Distributions of normalized latent weights.}
         \label{fig:discussion_distribution}
      \end{subfigure}
      \begin{subfigure}[t]{0.445\columnwidth}
         \centering
         \includegraphics[width=0.99\columnwidth]{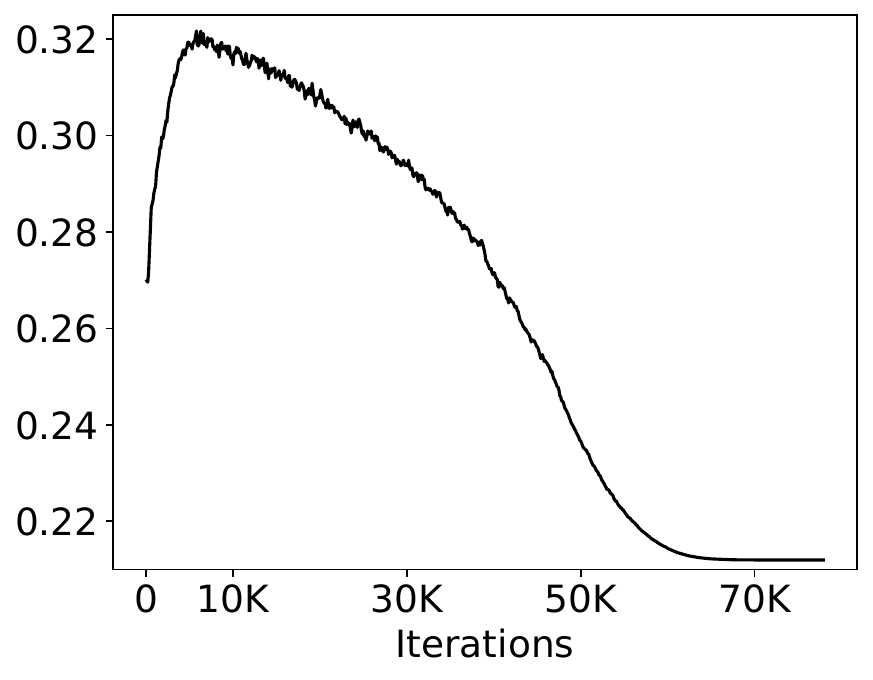}
         \caption{Average distances to the nearest transition points.}
         \label{fig:discussion_MD2TP}
      \end{subfigure}
   \end{center}
         \vspace{-0.5cm}
         \caption{Analysis on TR scheduling. We train ResNet-20~\cite{he2016deep} on CIFAR-100~\cite{krizhevsky2009learning} using SGDT, where we quantize both weights and activations with 2-bit representations. We visualize distributions of normalized latent weights in the 16$^\text{th}$ layer in~{\subref{fig:discussion_distribution}}, and average distances between normalized latent weights and the nearest transition points in~{\subref{fig:discussion_MD2TP}}. The transition points in~{\subref{fig:discussion_distribution}} are denoted by TPs in the x-axis. The top-1 test accuracy and average effective step sizes of quantized weights are shown by the red curves in Figs.~\ref{fig:teaser_acc} and~\ref{fig:teaser_quant}, respectively.} 
   \label{fig:discussion}
 \end{figure*}

\subsection{Experimental settings} \label{sec:settings}
For image classification, we train quantized models for MobileNetV2~\cite{sandler2018mobilenetv2}, ResNet families~\cite{he2016deep}, ReActNet-18~\cite{liu2020reactnet}, and DeiT-T/S~\cite{touvron2021training} on CIFAR-10/100~\cite{krizhevsky2009learning} and/or ImageNet~\cite{deng2009imagenet}. We train them using a cross-entropy loss, except for ReActNet-18 on ImageNet, where we use a distributional loss~\cite{liu2020reactnet} following the work of~\cite{liu2020reactnet}. For object detection, we adopt RetinaNet~\cite{lin2017focal} with ResNet backbones on MS COCO~\cite{lin2014microsoft}. Unlike the previous QAT methods~\cite{yamamoto2021learnable,zhuang2020training}, we use a shared prediction head to handle features of different resolutions, analogous to the original RetinaNet~\cite{lin2017focal}. For ease of activation quantization, we add a ReLU layer after each convolutional layer in the prediction head, so that all inputs of activation quantizers are non-negative. For more details, please refer to the Sec.~S4.1 of the supplement.

While our method requires additional computations~(\ie, element-wise comparison in Eq.~\eqref{eq:TR} and scalar operations in Eqs.~\eqref{eq:EMA_tr}-\eqref{eq:TALR}), they are computationally cheap compared to the whole training process. The training time increases by only 2\% compared to the plain optimization methods with the same machine~(Sec.~S2.5 of the supplement).

\subsection{Results} \label{sec:results}
\vspace{-0.1cm}
\paragraph{Image classification.} \label{sec:classification_results}
We provide in Tables~\ref{tab:imagenet_main}-\ref{tab:imagenet_vit} quantitative comparisons of quantized models trained with optimizers using plain optimization methods and our approach. We report a top-1 classification accuracy on ImageNet~\cite{deng2009imagenet} and CIFAR-100/10~\cite{krizhevsky2009learning} using the MobileNetV2~\cite{sandler2018mobilenetv2}, ReActNet-18~\cite{liu2020reactnet}, ResNet-18/20~\cite{he2016deep}, and DeiT-T/S~\cite{touvron2021training} architectures. From these tables, we observe three things: (1) Our method provides substantial accuracy gains over the plain optimizers, regardless of the datasets, network architectures, and quantization bit-widths. This indicates that scheduling a target TR is a better choice for the optimization process in QAT compared to the conventional strategy scheduling a LR. (2) The performance gaps on ImageNet using light-weight MobileNetV2~(0.6$\sim$6.7\%) are more significant than the ones using ReActNet-18 or ResNet-18~(0.1$\sim$0.5\%). Moreover, the performance gaps become larger for smaller bit-widths of MobileNetV2. These results suggest that the TR scheduling technique is especially useful for compressing networks aggressively, such as quantizing a light-weight model or extremely low-bit quantization. (3) Considering the results for ReActNet-18 and the ResNet families, our approach outperforms the conventional optimization methods by significant margins~(0.4$\sim$2.5\%) on the small dataset~(\ie, CIFAR-100/10). On the large-scale dataset~(\ie, ImageNet), it also shows superior results, achieving 0.1$\sim$0.5\% accuracy gains. The overall performance gaps decrease on ImageNet, possibly because the plain optimizers with a gradually decaying LR~(\eg, cosine annealing LR~\cite{loshchilov2016sgdr}) benefit from lots of training iterations on ImageNet~(roughly 600K). They, however, do not show satisfactory results within a small number of iterations on CIFAR-100/10~(roughly 80K), compared to ours.



\vspace{-0.45cm}
\paragraph{Object detection.}
We compare in Table~\ref{tab:ms_coco_main} the quantization performance of detection models in terms of an average precision~(AP) on the validation split of MS COCO~\cite{lin2014microsoft}. We train RetinaNet~\cite{lin2017focal} with the ResNet-50~\cite{he2016deep} backbone using either SGD or SGDT on the training split of MS COCO. We can observe in Table~\ref{tab:ms_coco_main} that the TR scheduling technique boosts the AP consistently over the SGD baselines across different bit-widths, similar to the results on image classification. This suggests that the TR scheduling technique is also useful for the object detection task involving both regression and classification, demonstrating once more the effectiveness of our method and its generalization ability to various tasks. Additional quantitative results on object detection with different backbone networks~(\eg, ResNet-18/34) and qualitative results are provided in the Sec.~S1.2 of the supplement.

\subsection{Analysis} \label{sec:analysis}
We show in Fig.~\ref{fig:discussion} an in-depth analysis on how a TR scheduler works during QAT. We can see from Fig.~\ref{fig:discussion_tr} that the running TR~$K^t$ roughly follows the target TR~$R^t$, indicating that we can control the average effective step size of quantized weights~(the red curve in Fig.~\ref{fig:teaser_quant}) by scheduling the target TR. This is possible because the TALR~$U^t$ is adjusted adaptively to match the running TR~$K^t$ with the target one~$R^t$~(Fig.~\ref{fig:discussion_talr}). We can see that the TALR~$U^t$ increases initially, since the running TR~$K^t$ is much smaller than the target TR~$R^t$. The TALR~$U^t$ then decreases gradually to reduce the number of transitions, following the target TR~$R^t$. Note that the TALR~$U^t$ approaches zero rapidly near the 50K-th iteration. To figure out the reason, we show in Figs.~\ref{fig:discussion_distribution} and~\ref{fig:discussion_MD2TP} distributions of normalized latent weights and their average distances to the nearest transition points, respectively. We can observe in Fig.~\ref{fig:discussion_distribution} that latent weights tend to be concentrated near the transition points of a quantizer as the training progresses, similar to the case in Sec.~\ref{sec:problem} using a user-defined LR. This implies that transitions occur more frequently in later training iterations if we do not properly reduce the degree of parameter change for latent weights. In particular, we can see in Fig.~\ref{fig:discussion_MD2TP} the average distances between the normalized latent weights and the nearest transition points are relatively small after the 50K-th iteration. Under such circumstance, the TALR should become much smaller in order to reduce the running TR, as in the sharp decline around the 50K-th iteration. We can thus conclude that our approach adjusts the TALR by considering the distribution of the latent weights implicitly.

\vspace{-0.15cm}
\section{Conclusion}
\vspace{-0.15cm}
We have discussed the problem of a conventional optimization method using a LR in QAT. To overcome this, we have presented a TR scheduling technique specialized for general QAT, which controls the number of transitions in quantized weights explicitly. We have shown that the optimizers coupled with our TR scheduling technique outperform the plain ones using a LR by significant margins under various QAT settings. We have also verified that our approach enables stable training with different types of optimizers and schedulers, which is generally difficult in the conventional optimization methods, indicating that the TR scheduling technique offers more diverse training options for QAT. We expect that our method would be adopted to boost the performance of other quantized models, bringing a significant advance in the field of network quantization.

\section*{Acknowledgements}
This work was supported in part by NRF and IITP grants funded by the Korea government (MSIT) (No. 2023R1A2C2004306, No.RS-2022-00143524, Development of Fundamental Technology and Integrated Solution for Next-Generation Automatic Artificial Intelligence System, RS-2021-II212068, Artificial Intelligence Innovation Hub) and the Yonsei Signature Research Cluster Program of 2025 (2025-22-0013).



{
    \small
    \bibliographystyle{ieeenat_fullname}
    \bibliography{main}
}

\clearpage
\newpage

\setcounter{equation}{0}
\setcounter{figure}{0}
\setcounter{table}{0}
\setcounter{section}{0}
\setcounter{subsection}{0}
\setcounter{subsubsection}{0}

\renewcommand{\theequation}{S\arabic{equation}}
\renewcommand{\thefigure}{S\arabic{figure}}
\renewcommand{\thetable}{S\arabic{table}}
\renewcommand{\thesection}{S\arabic{section}}
\renewcommand{\thesubsection}{S\arabic{section}.\arabic{subsection}}
\renewcommand{\thesubsubsection}{S\arabic{section}.\arabic{subsection}.\arabic{subsubsection}}

\newcolumntype{L}[1]{>{\raggedright\let\newline\\\arraybackslash\hspace{0pt}}m{#1}} 
\newcolumntype{R}[1]{>{\raggedleft\let\newline\\\arraybackslash\hspace{0pt}}m{#1}}
\newcolumntype{C}[1]{>{\centering\let\newline\\\arraybackslash\hspace{0pt}}m{#1}}


\maketitlesupplementary


In the supplement, we give additional results in image classification and object detection~(Sec.~\ref{sec:appendix_results}). We then provide in-depth analyses~(Sec.~\ref{sec:appendix_analysis}) and discussions on our method~(Sec.~\ref{sec:appendix_discussion}). Finally, we present the implementation details of our method~(Sec.~\ref{sec:appendix_implementation}) and comprehensive description of a quantizer~(Sec.~\ref{sec:appendix_supp_quantizer}). We summarize in Algorithm~\ref{alg:algorithm} an overall process of our method.


\section{Results}
\label{sec:appendix_results}
\subsection{Image classification}
\paragraph{Comparison to the state of the art}
We present in Table~\ref{tab:appendix_sota} a quantitative comparison of our approach and state-of-the-art methods for QAT~\cite{choi2018pact,jung2019learning,yang2019quantization,gong2019differentiable,esser2019learned,bhalgat2020lsq+,lee2021network,nagel2022overcoming} on ImageNet~\cite{deng2009imagenet}, in terms of a top-1 validation accuracy.  For a fair comparison, we perform comparisons with the methods using uniform quantization schemes and a vanilla architecture. We can see from the table that our method~(\ie, using either SGDT or AdamT) achieves state-of-the-art performance for ResNet-18 under the 1/1 and 2/2 bit-width settings, and shows competitive performance for the 3/3 and 4/4 bit-width settings. In terms of performance gains or drops compared to the full-precision models, our results are better than or on a par with the state of the art. For MobileNetV2, our method also sets a new state of the art, except for the 3/3 bit-width setting. The work of~\cite{nagel2022overcoming} alleviates the oscillation problem in QAT by freezing latent weights or adding a regularization term. It adopts architecture-specific and bit-specific hyperparameters, such as a LR, a weight decay, and a threshold for weight freezing. This is because freezing weights and adding the regularizer could potentially degrade trainability and disturb the training process, making the QAT process sensitive to hyperparameters. In contrast, our method is more practical as it performs consistently well under the various bit-width settings with the same set of hyperparameters. For completeness, we also try to tune a TR factor~$\lambda$ specific to the 3/3-bit setting of MobileNetV2, and improve the performance with the TR factor~$\lambda$ of 2e-3.

Other QAT methods mostly focus on designing a quantizer in a forward step~(\eg,~PACT~\cite{choi2018pact}, QIL~\cite{jung2019learning}, LSQ~\cite{esser2019learned}, LSQ+~\cite{bhalgat2020lsq+}) or addressing a gradient mismatch problem in a backward step~(\eg,~QNet~\cite{yang2019quantization}, DSQ~\cite{gong2019differentiable}, EWGS~\cite{lee2021network}). Our approach is orthogonal to these methods in that we focus on an optimization step in QAT. Namely, we propose to schedule a target TR instead of a LR itself, and introduce a novel TALR to update latent weights. Compared to other methods that require computationally expensive operations for differentiable quantizers~(\eg,~QNet\cite{yang2019quantization},~DSQ~\cite{gong2019differentiable}) or approximating a hessian trace~(\eg, EWGS~\cite{lee2021network}), ours is relatively simple and efficient, since it adds few comparison and scalar operations only in the QAT process.

\begin{algorithm}[t]
   \small
   \caption{Optimization process using a TR scheduler.} \label{alg:algorithm}
   \textbf{Require}: the~number~of~iterations~$T$; a target~TR~$R^t$; a momentum constant~$m$.
   \begin{algorithmic}[1]
      \STATE {\textbf{while} $t < T$ \textbf{do}}
      \STATE ~~Compute a current TR~$k^t$ [Eq.~(5)]:\\
      ~~~$k^t \leftarrow \frac{\sum^{N}_{i=1} \mathbb{I}\left[ w^t_d (i) \neq w^{t-1}_d (i) \right]}{N}$.\\
      \STATE ~~Estimate a running TR~$K^t$ [Eq.~(10)]:\\
      ~~~$K^t \leftarrow m K^{t-1} + (1-m)k^t$.\\
      \STATE ~~Adjust a TALR~$U^t$ [Eq.~(11)]:\\
      ~~~$U^t \leftarrow \max \left( 0, U^{t-1} + \eta \left( R^t - K^t \right) \right)$.\\
      \STATE ~~Update latent weights~${\bf{w}}^t$ with a gradient term~${\bf{g}}^{t}$ \\~~and the TALR~$U^t$ [Eq.~(12)]:\\
      ~~~${\bf{w}}^{t+1} \leftarrow {\bf{w}}^t - U^t {\bf{g}}^{t}$.\\
      \STATE {\textbf{end while}}
   \end{algorithmic}
 \end{algorithm}

\begin{table*}[t]
   \setlength{\tabcolsep}{0.3em}
   \centering
   \small
   \caption{Quantitative comparison of our method and the state of the art on ImageNet~\cite{deng2009imagenet} in terms of a top-1 validation accuracy. The numbers in parentheses indicate the performance gains or drops compared to the full-precision~(32/32) models. All numbers are taken from corresponding works except for LSQ~\cite{esser2019learned}, where the results are reproduced in~\cite{bhalgat2020lsq+} using a vanilla structure of ResNet~\cite{he2016deep}.}
   \vspace{-0.1cm}
      \begin{tabular}{C{2.4cm} L{2.6cm} C{1.7cm} C{1.8cm} C{1.8cm} C{1.8cm} C{1.8cm} C{1.3cm}}
         \midrule
         \multirow{2}{*}{Architecture} & \multicolumn{1}{c}{\multirow{2}{*}{Methods}} & \multirow{2}{*}{Optimizer} & \multicolumn{5}{c}{Weight/activation bit-widths} \\
                                       &                                               &                            & 1/1 & 2/2 & 3/3 & 4/4 & 32/32 \\
         \midrule 
         \multirow{9}{*}{ResNet-18}  & PACT~\cite{choi2018pact}          & SGD   & -                                      & 64.4~($-$5.8)                         & 68.1~($-$2.1)                         & 69.2~($-$1.0) & 70.2 \\
                                     & QIL~\cite{jung2019learning}       & SGD   & -                                      & 65.7~($-$4.5)                         & 69.2~($-$1.0)                         & 70.1~($-$0.1) & 70.2 \\
                                     & QNet~\cite{yang2019quantization}  & SGD   & 53.6~($-$16.7)                         & -                                     & -                                     & -             & 70.3 \\
                                     & DSQ~\cite{gong2019differentiable} & -     & -                                      & 65.2~($-$4.7)                         & 68.7~($-$1.2)                         & 69.6~($-$0.3) & 69.9 \\
                                     & LSQ~\cite{esser2019learned}       & SGD   & -                                      & 66.7~($-$3.4)                         & \underline{69.4}~(\underline{$-$0.7}) & \underline{70.7}~(\underline{$+$0.6}) & 70.1 \\
                                     & LSQ+~\cite{bhalgat2020lsq+}       & SGD   & -                                      & 66.8~($-$3.3)                         & 69.3~($-$0.8)                         & {\bf{70.8}}~({\bf{$+$0.7}})      & 70.1 \\
                                     & EWGS~\cite{lee2021network}        & SGD   & 55.3~($-$14.6)                         & \underline{67.0}~(\underline{$-$2.9}) & {\bf{69.7}}~({\bf{$-$0.2}})           & 70.6~({\bf{$+$0.7}}) & 69.9 \\
                                     & Ours                              & SGDT  & \underline{55.8}~(\underline{$-$14.1}) & 66.9~($-$3.0)                         & {\bf{69.7}}~({\bf{$-$0.2}})           & 70.6~({\bf{$+$0.7}}) & 69.9 \\
                                     & Ours                              & AdamT & {\bf{56.3}}~({\bf{$-$13.6}})           & {\bf{67.2}}~({\bf{$-$2.7}})           & {\bf{69.7}}~({\bf{$-$0.2}})           & 70.4~($+$0.5)                    & 69.9 \\
         \midrule
         \multirow{6}{*}{MobileNetV2}  & DSQ~\cite{gong2019differentiable}      & -     & - & -                                      & -                                & 64.8~($-$7.1)                         & 71.9 \\
                                       & EWGS~\cite{lee2021network}             & SGD   & - & -                                      & -                                & 70.3~($-$1.6)                         & 71.9 \\
                                       & Nagel~\etal~\cite{nagel2022overcoming} & SGD   & - & -                                      & \underline{67.6}~({\bf{$-$4.1}}) & \underline{70.6}~({\bf{$-$1.1}})      & 71.7 \\
                                       & Ours                                   & SGDT  & - & \underline{53.6}~(\underline{$-$18.3}) & 67.0~($-$4.9)                    & 70.5~(\underline{$-$1.4})             & 71.9 \\
                                       & Ours                                   & AdamT & - & {\bf{53.8}}~({\bf{$-$18.1}})           & 67.3~(\underline{$-$4.6})        & {\bf{70.8}}~({\bf{$-$1.1}})           & 71.9 \\
                                       & Ours~($\lambda$=2e-3)                  & AdamT & - & -                                      & {\bf{67.8}}~({\bf{$-$4.1}})      & -                                     & 71.9 \\
         \midrule
      \end{tabular} \label{tab:appendix_sota}
 \end{table*}

\begin{figure*}[t]
   \captionsetup[subfigure]{justification=centering}
   \begin{center}
      \begin{subfigure}[t]{0.535\columnwidth}
         \centering
         \includegraphics[width=1\columnwidth]{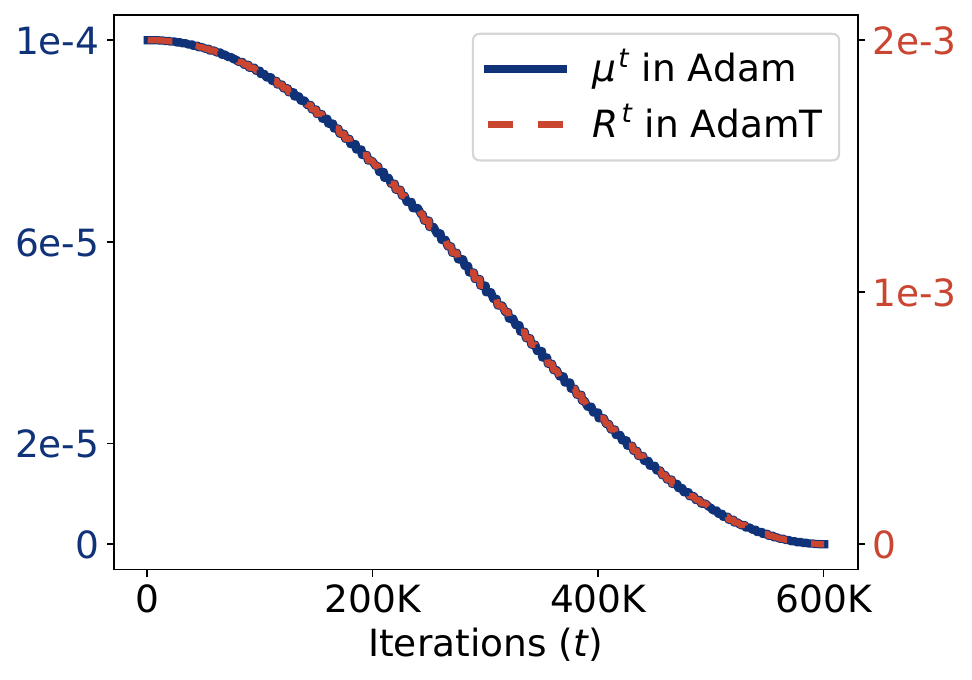}
      \end{subfigure}
      \begin{subfigure}[t]{0.508\columnwidth}
         \centering
         \includegraphics[width=1\columnwidth,right]{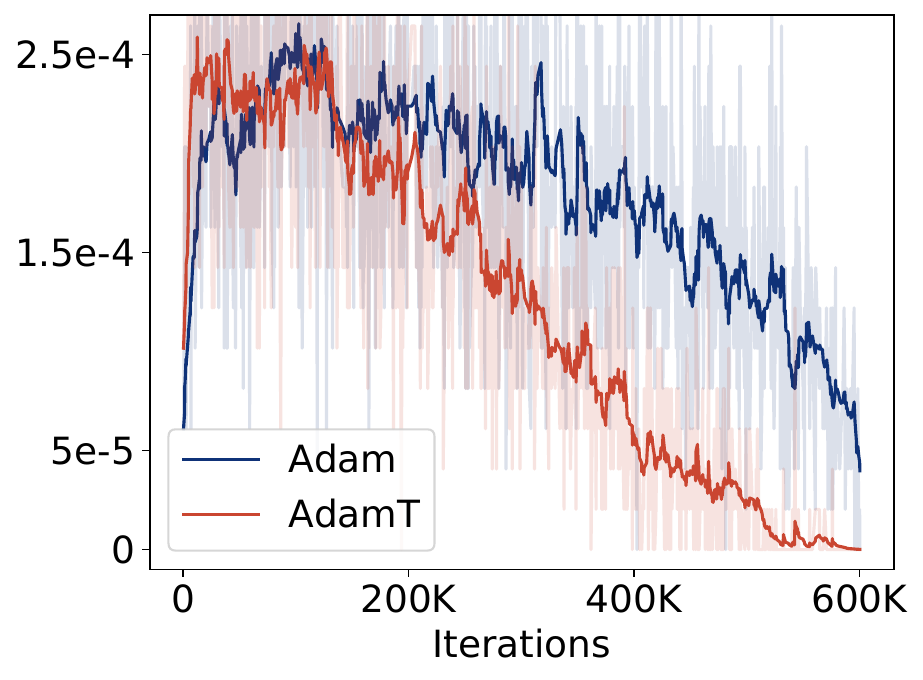} 
      \end{subfigure}
      \begin{subfigure}[t]{0.475\columnwidth}
         \centering
         \includegraphics[width=1\columnwidth]{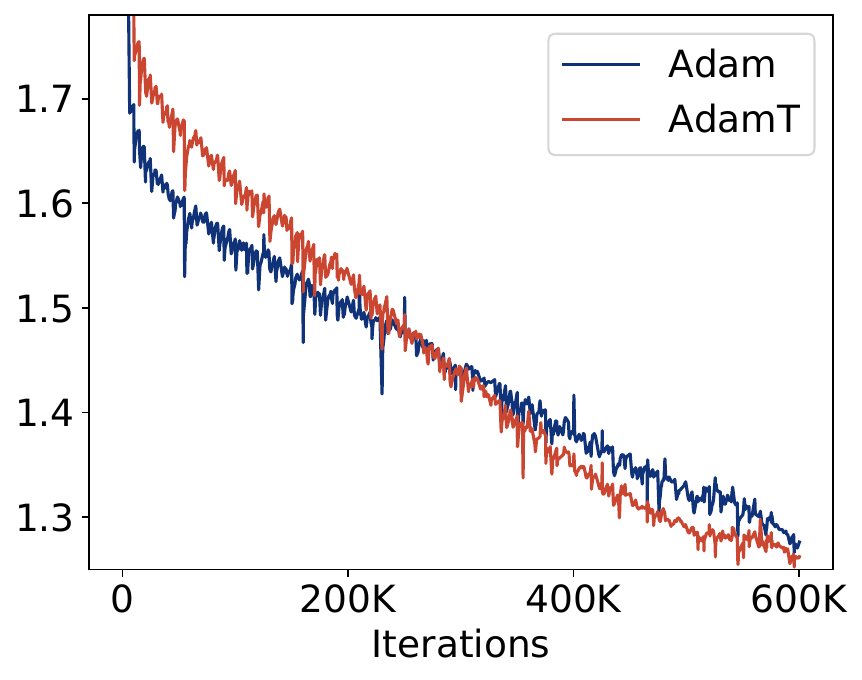}
      \end{subfigure}
      \begin{subfigure}[t]{0.465\columnwidth}
         \centering
         \includegraphics[width=1\columnwidth]{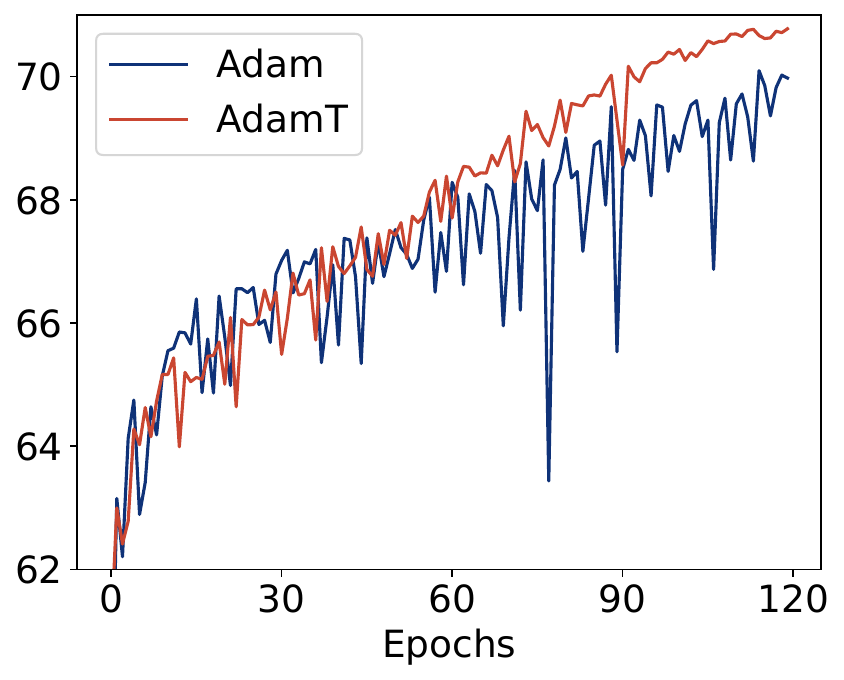}
      \end{subfigure}
   \end{center}

   \begin{center}
      \begin{subfigure}[t]{0.535\columnwidth}
         \centering
         \includegraphics[width=1\columnwidth]{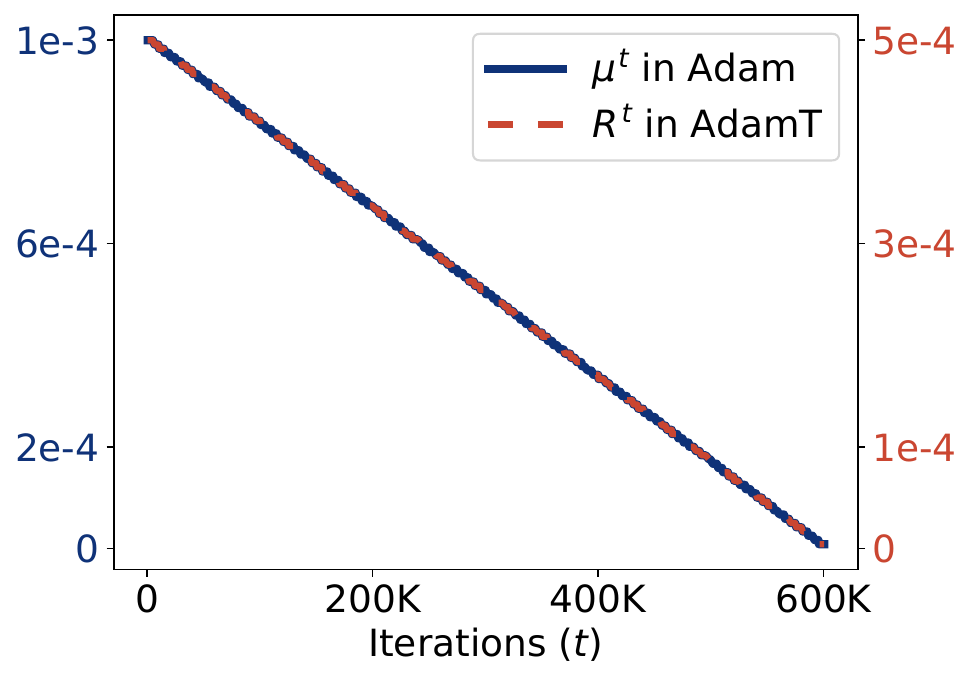}
         \caption{LR~$\mu$ and target TR~$R^t$.}
         \label{fig:appendix_training_curves_lr_tr}
      \end{subfigure}
      \begin{subfigure}[t]{0.508\columnwidth}
         \centering
         \includegraphics[width=0.96\columnwidth,right]{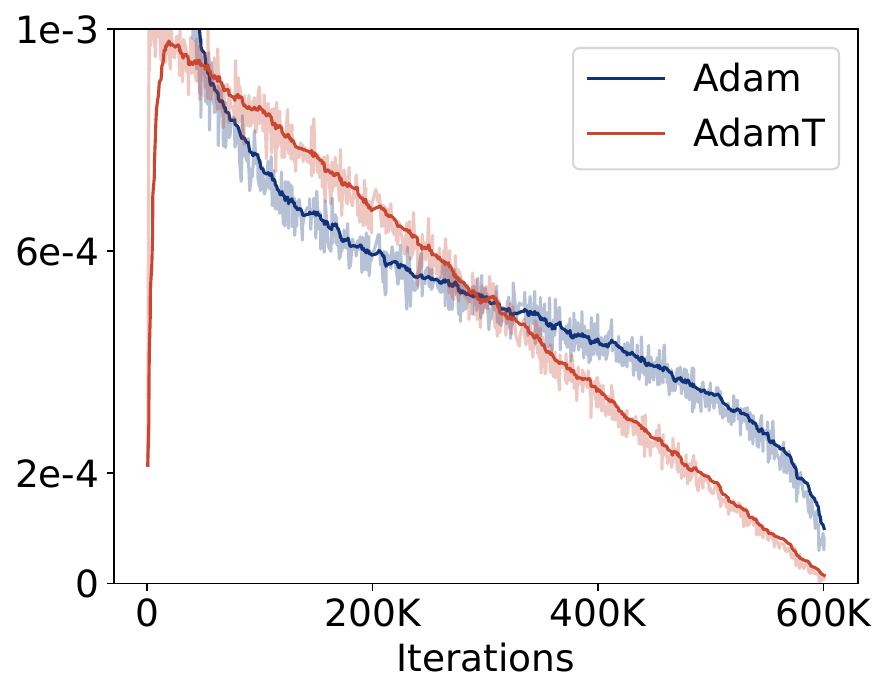} 
         \caption{Avg. effective step \newline sizes of quantized weights.}
         \label{fig:appendix_training_curves_stepsize}
      \end{subfigure}
      \begin{subfigure}[t]{0.475\columnwidth}
         \centering
         \includegraphics[width=1\columnwidth]{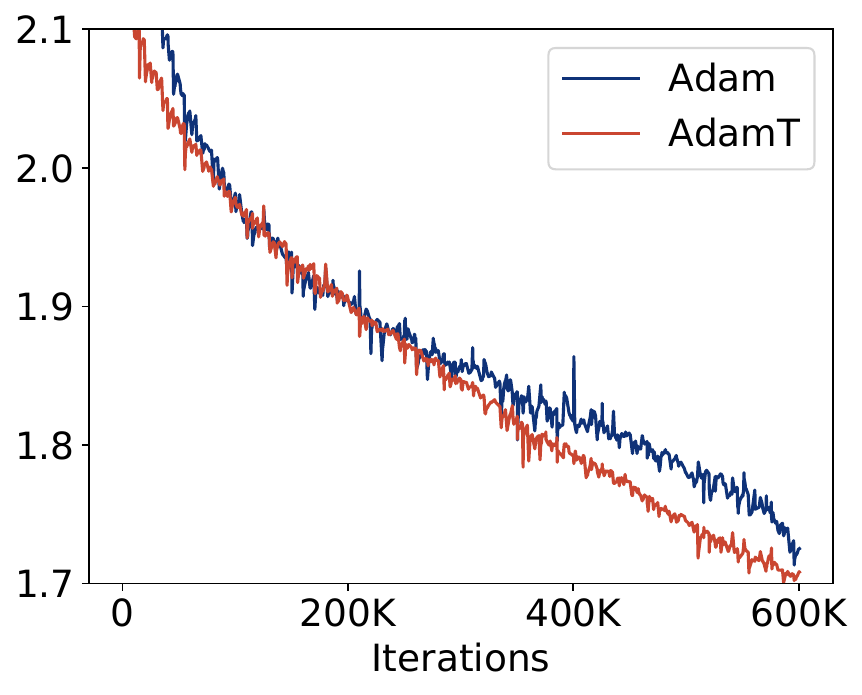}
         \caption{Running avg. of \newline training losses.}
         \label{fig:appendix_training_curves_loss}
      \end{subfigure}
      \begin{subfigure}[t]{0.465\columnwidth}
         \centering
         \includegraphics[width=1\columnwidth]{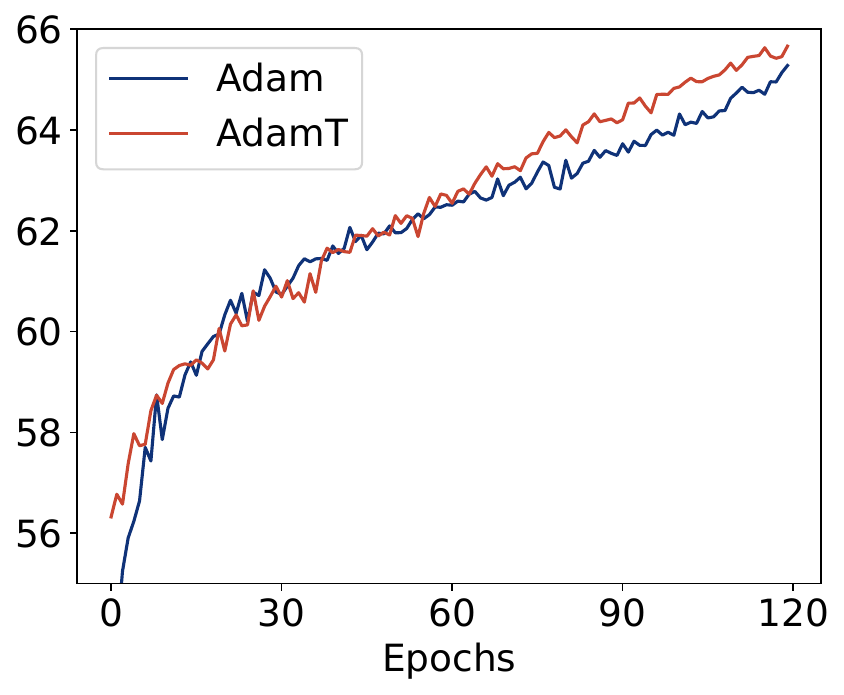}
         \caption{Top-1 validation accuracy.}
         \label{fig:appendix_training_curves_acc}
      \end{subfigure}
   \end{center}
         \vspace{-0.1cm}
         \caption{Training curves for quantized models using Adam~\cite{kingma2014adam} and AdamT on ImageNet~\cite{deng2009imagenet}. The results in the first and second rows are obtained with MobileNetV2~\cite{sandler2018mobilenetv2} and ReActNet-18~\cite{liu2020reactnet} using 4-bit and binary weights/activations, respectively. For the visualizations in~{\subref{fig:appendix_training_curves_stepsize}}, we monitor the average effective step sizes of quantized weights in the 19$^\text{th}$ and 17$^\text{th}$ layers of MobileNetV2 and ReActNet-18, respectively. (Best viewed in color.)}
   \label{fig:appendix_training_curves}
 \end{figure*}

\begin{table*}[t]
   \setlength{\tabcolsep}{0.3em}
   \centering
   \small
   \caption{Quantitative comparison of quantized models on object detection. We train RetinaNet~\cite{lin2017focal} on the training split of MS COCO~\cite{lin2014microsoft} with different backbone networks and quantization bit-widths, using either the plain optimization method~(SGD) or ours~(SGDT). We report the average precision~(AP) on the validation split.}
   \vspace{-0.1cm}
     \begin{tabular}{C{2.3cm}  C{1.2cm}  C{1.7cm}  C{1.2cm} C{1.2cm} C{1.2cm} C{1.2cm} C{1.2cm} C{1.2cm}}
       \midrule
       Backbone                         & W/A                   & Optimizer & AP         & AP$_{50}$  & AP$_{75}$  & AP$_{S}$   & AP$_{M}$   & AP$_{L}$ \\
       \midrule 

       \multirow{5}{*}[-3pt]{ResNet-34} & FP                    & SGD       & 37.70      & 57.32      & 40.71      & 22.35      & 41.50      & 48.93 \\ \cmidrule{2-9}
                                        & \multirow{2}{*}{4/4}  & SGD       & 37.99      & 57.29      & 40.34      & \bf{22.96} & 41.16      & 49.32 \\
                                        &                       & SGDT      & \bf{38.14} & \bf{57.39} & \bf{40.93} & 22.05      & \bf{41.70} & \bf{49.79} \\ \cmidrule{2-9}
                                        & \multirow{2}{*}{3/3}  & SGD       & 37.32      & 56.61      & 39.95      & \bf{22.22} & \bf{40.54} & 49.16 \\
                                        &                       & SGDT      & \bf{37.64} & \bf{56.78} & \bf{40.28} & 21.81      & 40.42      & \bf{49.89} \\ \cmidrule{1-9}
       \multirow{5}{*}[-3pt]{ResNet-18} & FP                    & SGD       & 34.06      & 53.15      & 36.17      & 19.66      & 36.48      & 44.45 \\ \cmidrule{2-9}
                                        & \multirow{2}{*}{4/4}  & SGD       & 35.09      & 54.05      & 37.40      & \bf{20.34} & \bf{37.62} & 46.69 \\
                                        &                       & SGDT      & \bf{35.30} & \bf{54.46} & \bf{37.59} & 19.92      & 37.61      & \bf{46.83} \\ \cmidrule{2-9}
                                        & \multirow{2}{*}{3/3}  & SGD       & 34.32      & 53.09      & 36.44      & 19.34      & 36.46      & 45.33 \\
                                        &                       & SGDT      & \bf{34.72} & \bf{53.63} & \bf{36.84} & \bf{19.91} & \bf{37.21} & \bf{45.64} \\
 
       \midrule
     \end{tabular} \label{tab:appendix_ms_coco_main}
 \end{table*}

\begin{figure*}[t]
   \captionsetup[subfigure]{justification=centering}
   \centering
      \begin{subfigure}{0.161\textwidth}
         \includegraphics[width=\textwidth,height=0.708\textwidth,frame]{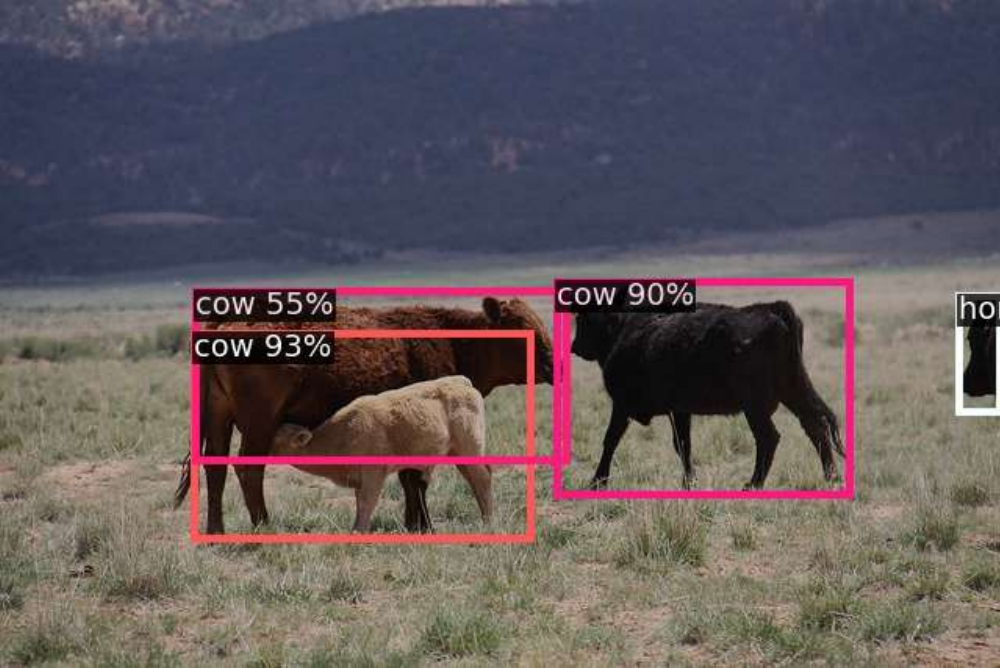}
      \end{subfigure}
      \begin{subfigure}{0.161\textwidth}
         \includegraphics[width=\textwidth,height=0.708\textwidth,frame]{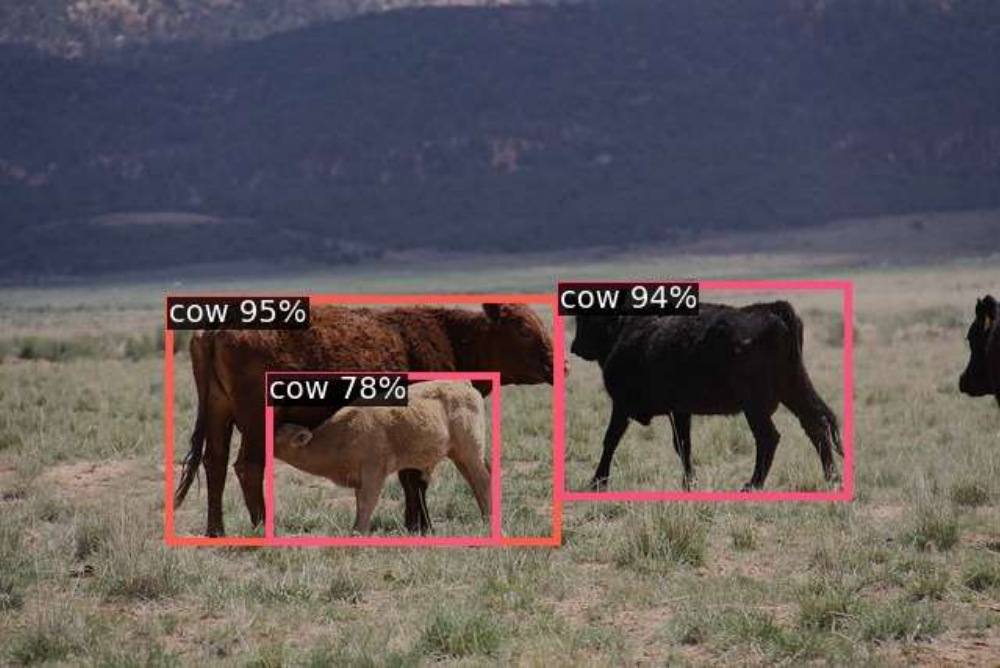}
      \end{subfigure}
      \begin{subfigure}{0.161\textwidth}
         \includegraphics[width=\textwidth,height=0.708\textwidth,frame]{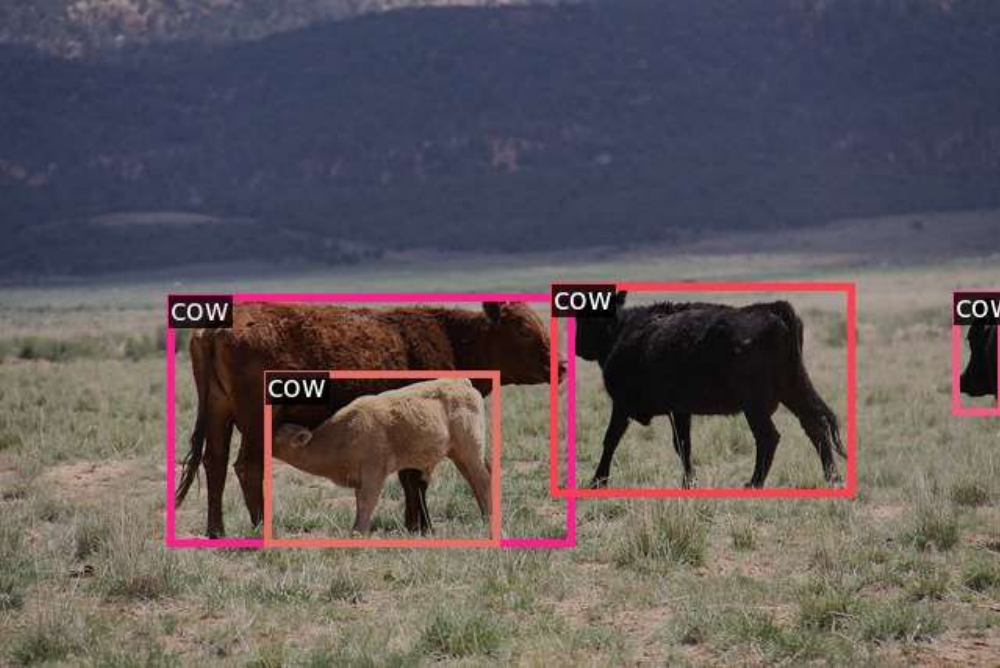}
      \end{subfigure}
      \hspace{0.05cm}
      \begin{subfigure}{0.161\textwidth}
         \includegraphics[width=\textwidth,height=0.708\textwidth,frame]{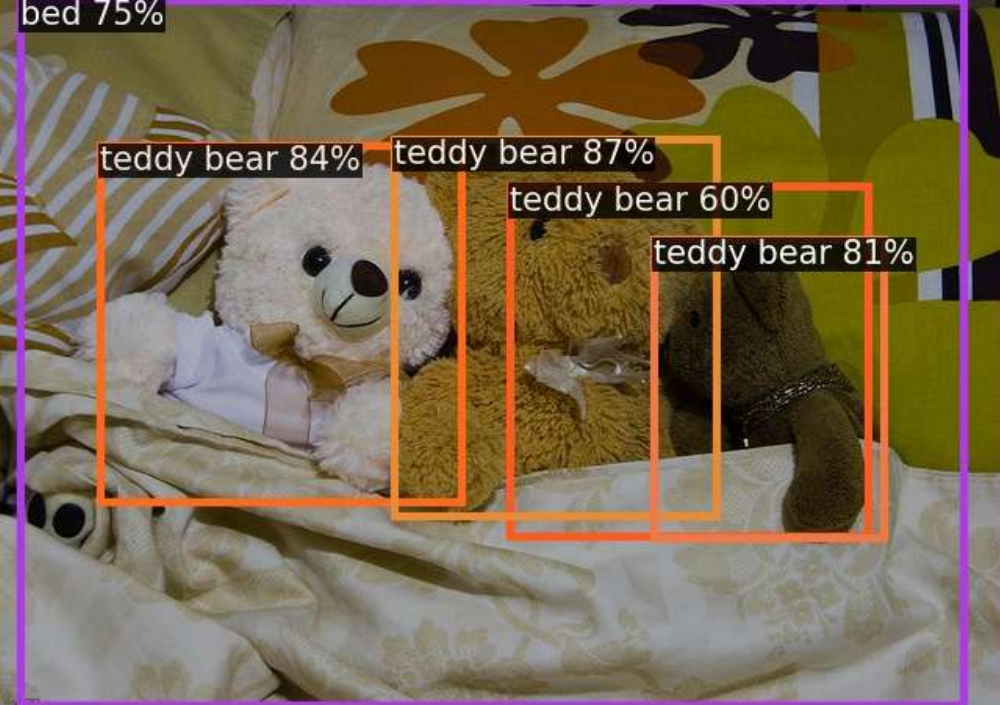}
      \end{subfigure}
      \begin{subfigure}{0.161\textwidth}
         \includegraphics[width=\textwidth,height=0.708\textwidth,frame]{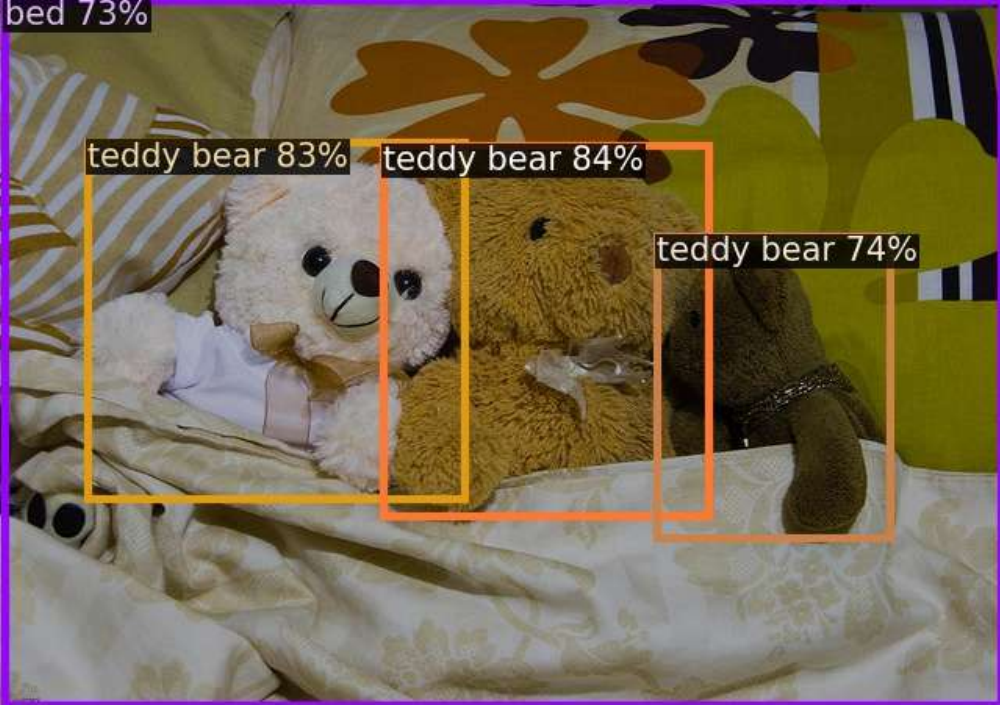}
      \end{subfigure}
      \begin{subfigure}{0.161\textwidth}
         \includegraphics[width=\textwidth,height=0.708\textwidth,frame]{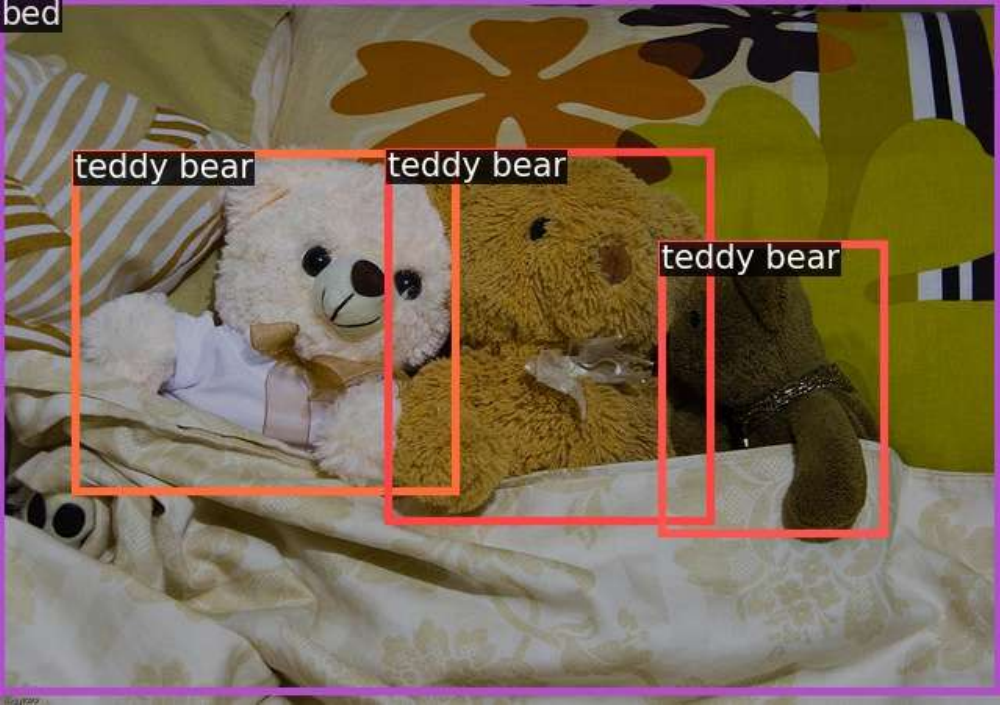}
      \end{subfigure}

      \vspace{0.1cm}

      \begin{subfigure}{0.161\textwidth}
         \includegraphics[width=\textwidth,height=0.708\textwidth,frame]{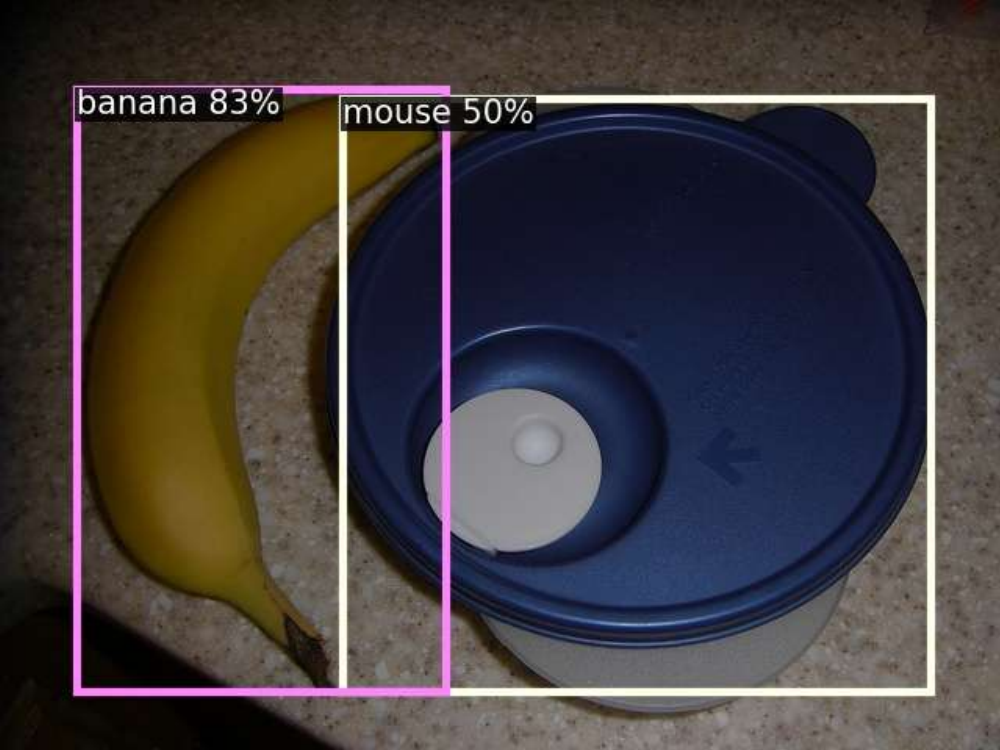}
         \caption*{Baseline.}
      \end{subfigure}
      \begin{subfigure}{0.161\textwidth}
         \includegraphics[width=\textwidth,height=0.708\textwidth,frame]{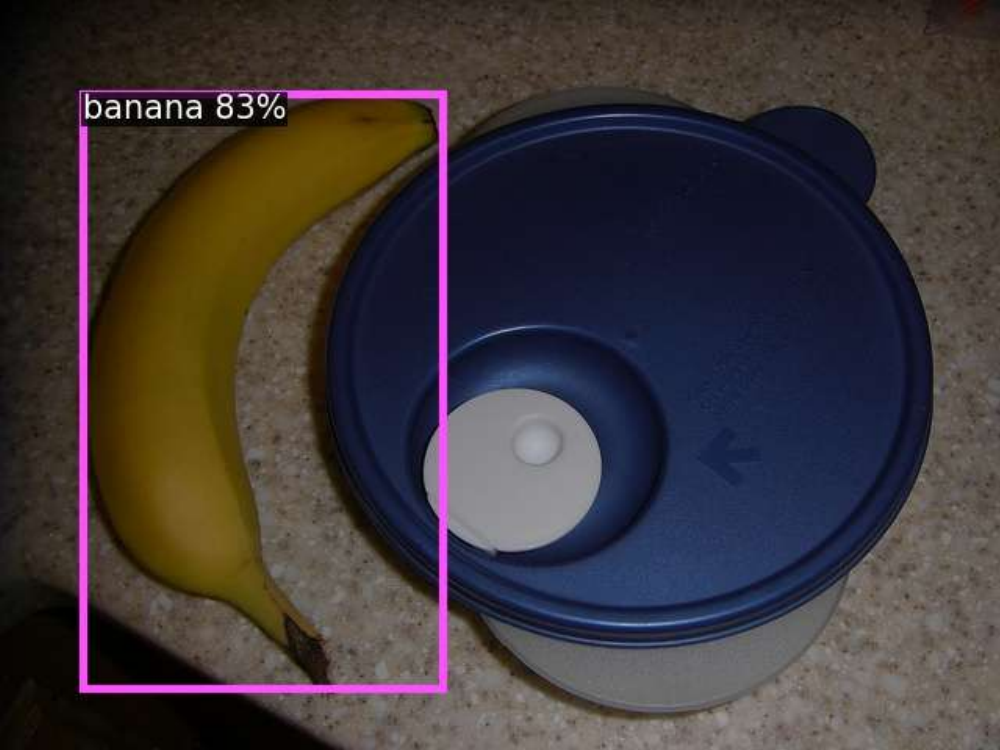}
         \caption*{Ours.}
      \end{subfigure}
      \begin{subfigure}{0.161\textwidth}
         \includegraphics[width=\textwidth,height=0.708\textwidth,frame]{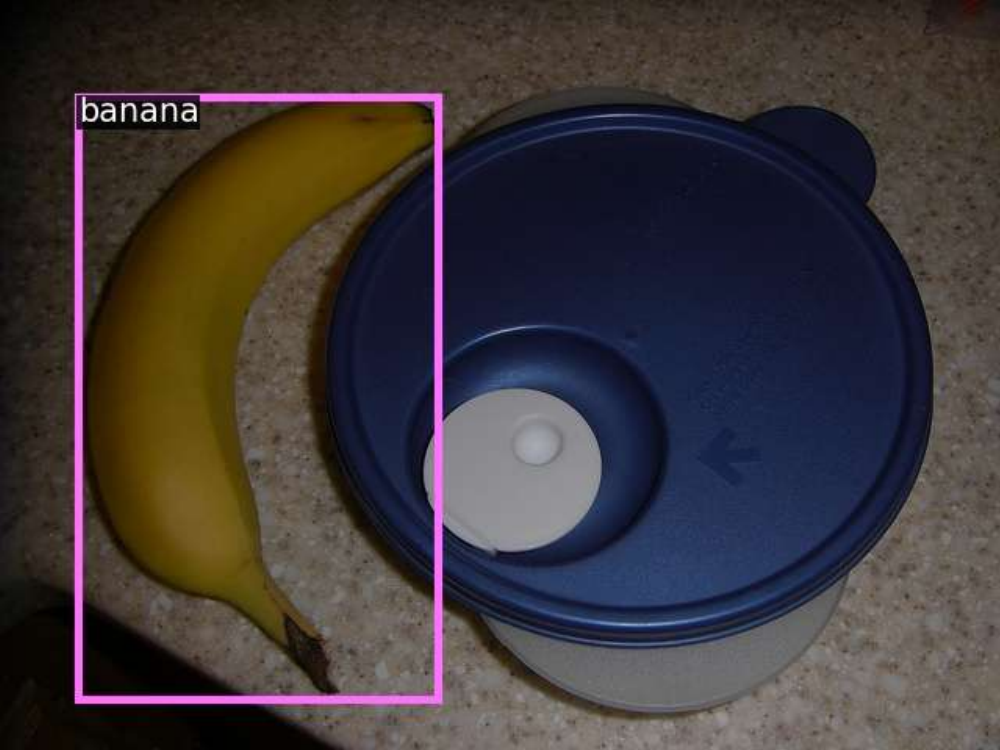}
         \caption*{Ground truth.}
      \end{subfigure}
      \hspace{0.05cm}
      \begin{subfigure}{0.161\textwidth}
         \includegraphics[width=\textwidth,height=0.708\textwidth,frame]{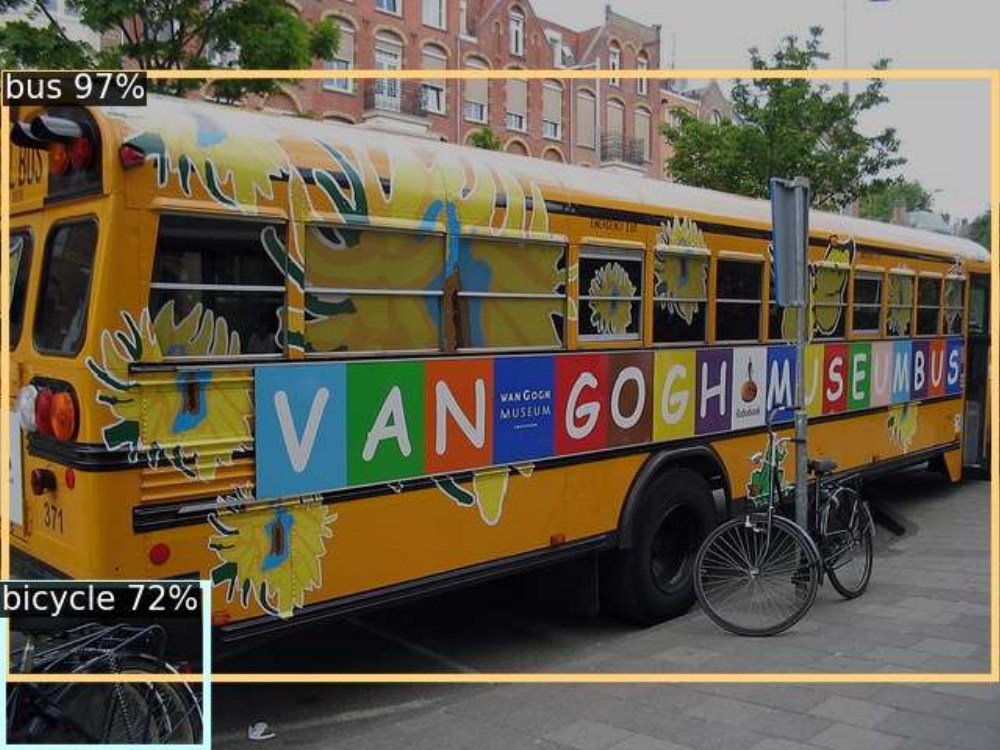}
         \caption*{Baseline.}
      \end{subfigure}
      \begin{subfigure}{0.161\textwidth}
         \includegraphics[width=\textwidth,height=0.708\textwidth,frame]{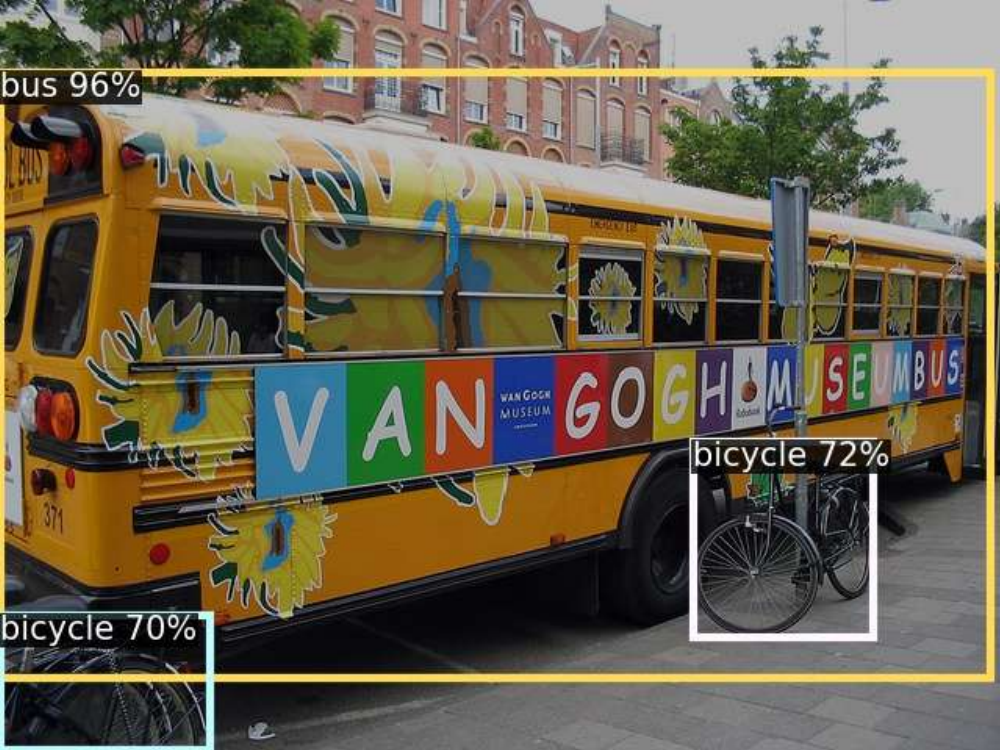}
         \caption*{Ours.}
      \end{subfigure}
      \begin{subfigure}{0.161\textwidth}
         \includegraphics[width=\textwidth,height=0.708\textwidth,frame]{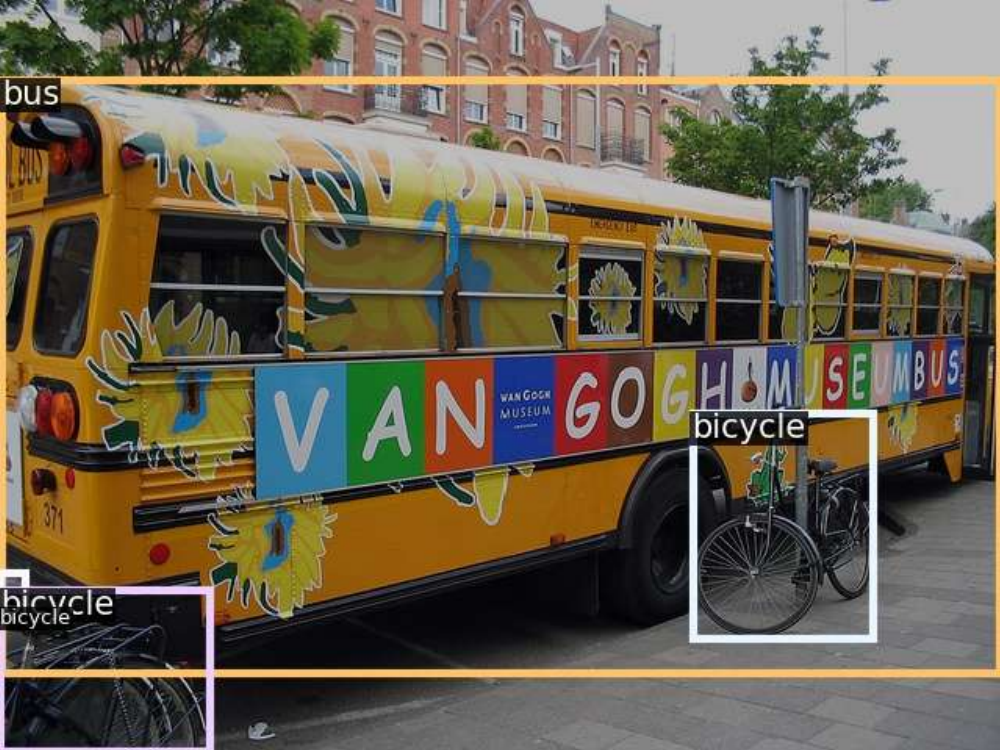}
         \caption*{Ground truth.}
      \end{subfigure}
   \vspace{-0.2cm}
   \caption{Qualitative comparison for object detection on MS COCO~\cite{lin2014microsoft} using RetinaNet~\cite{lin2017focal} with the ResNet-50~\cite{he2016deep} backbone, where both weights and activations are quantized into 4-bit. The results of baseline and ours are obtained from the models trained with SGD and SGDT, respectively. (Best viewed in color.)}
   \label{fig:appendix_ms_coco}	
\end{figure*}

\paragraph{Quantitative results.}
We show in Fig.~\ref{fig:appendix_training_curves} training curves on ImageNet for MobileNetV2 and ReActNet-18 using 4-bit and binary weights/activations, respectively. We can see from Figs.~\ref{fig:appendix_training_curves_lr_tr} and~\ref{fig:appendix_training_curves_stepsize} that the optimizer coupled with our TR scheduling technique~(\ie, AdamT) can control the average effective step size of quantized weights roughly using a target TR. On the other hand, the plain optimizer adopting a user-defined LR~(\ie, Adam) fails to control it by scheduling the LR. For example, the average effective step sizes for quantized weights in MobileNetV2 (\ie, the blue curve in the first row of Fig.~\ref{fig:appendix_training_curves_stepsize}) do not approach zero at the end of training. Such large changes in the quantized weights can lead to unstable batch normalization statistics~\cite{ioffe2015batch}, resulting in noisy and degraded test time performance~\cite{nagel2022overcoming,park2020profit} (\ie, the blue curve in the first row of Fig.~\ref{fig:appendix_training_curves_acc}). On the contrary, the TR scheduling technique enables better training in terms of the convergence rate and performance, which can be verified by the training losses and validation accuracy in Figs.~\ref{fig:appendix_training_curves_loss} and~\ref{fig:appendix_training_curves_acc}, respectively.

\subsection{Object detection}
\paragraph{Quantitative results.}
We compare in Table~\ref{tab:appendix_ms_coco_main} the quantization performance of detection models in terms of an average precision~(AP) on the validation split of MS COCO~\cite{lin2014microsoft}. We train RetinaNet~\cite{lin2017focal} with the ResNet-18/34~\cite{he2016deep} backbones using either SGD or SGDT on the training split of MS COCO. From Table~4 of the main paper and Table~\ref{tab:appendix_ms_coco_main}, we can see that using a TR scheduling technique improves the performance of quantized models in terms of AP, compared to the plain optimization method, across different bit-width and backbone settings. This confirms once again the effectiveness and generalization ability of our approach.

\paragraph{Qualitative results.}
We compare in Fig.~\ref{fig:appendix_ms_coco} qualitative results of baseline~(SGD) and our~(SGDT) models, using RetinaNet with the ResNet-50 backbone, with 4-bit weights/activations. The baseline model trained with SGD provides incorrect bounding boxes in the presence of overlapped/adjacent instances~(\eg, the cow and teddy bear in the first and second examples, respectively), misclassifies object classes~(\eg, the food container in the third example), and fails to detect objects~(\eg, the bicycle in the last example). In contrast, the model trained with our approach offers accurate bounding boxes, and classifies object classes successfully.

\begin{figure*}[t]
   \captionsetup[subfigure]{justification=centering}
   \begin{center}
      \begin{subfigure}[t]{0.495\columnwidth}
         \centering
         \includegraphics[width=1\columnwidth]{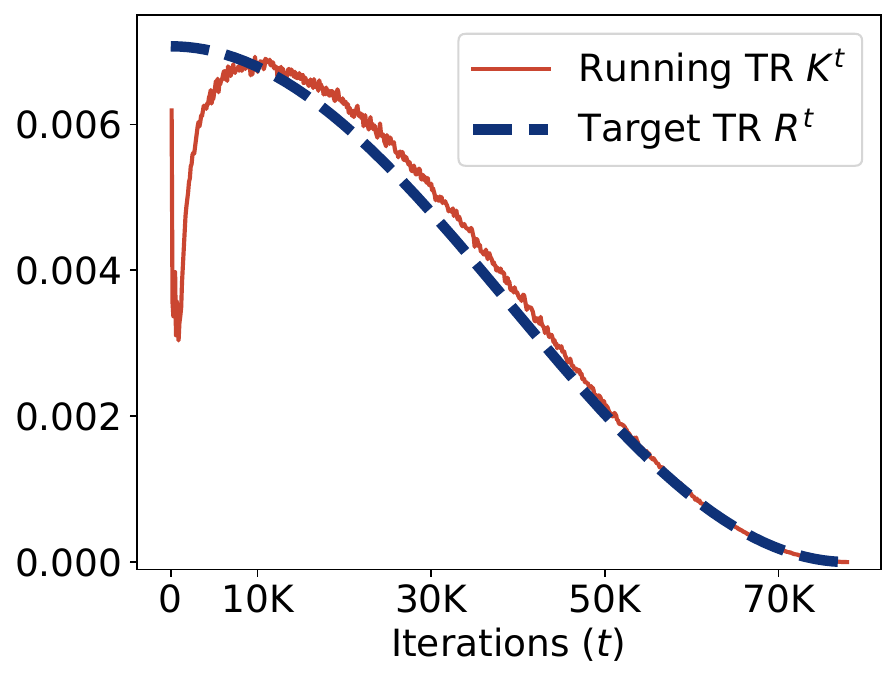}
         \caption{Running TR~$K^t$ and \newline target TR~$R^t$.}
         \label{fig:appendix_discussion_tr}
      \end{subfigure}
      \begin{subfigure}[t]{0.468\columnwidth}
         \centering
         \includegraphics[width=1\columnwidth]{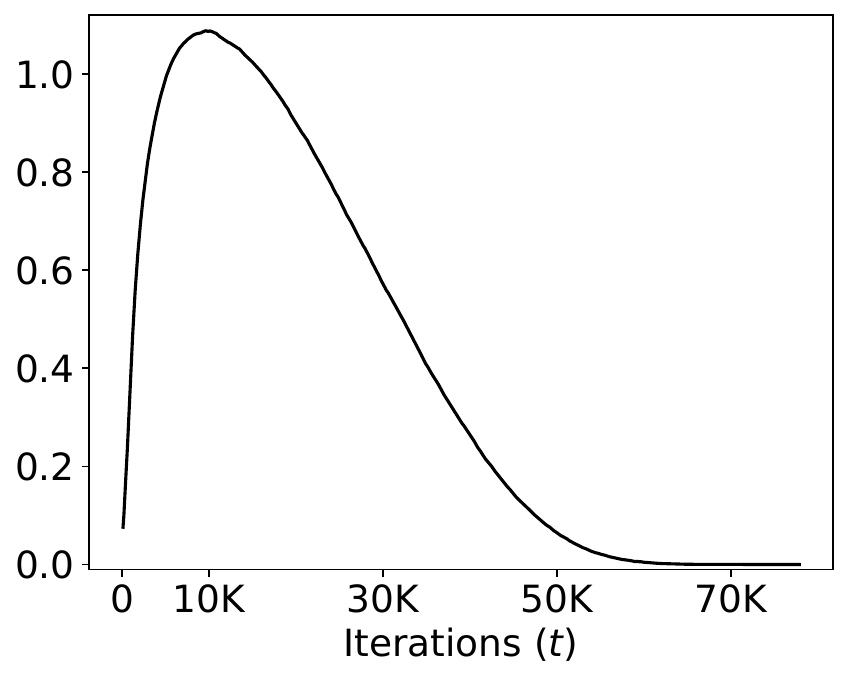}
         \caption{TALR~$U^t$.}
         \label{fig:appendix_discussion_talr}
      \end{subfigure}
      \begin{subfigure}[t]{0.495\columnwidth}
         \centering
         \raisebox{0.148cm}{\includegraphics[width=1\columnwidth]{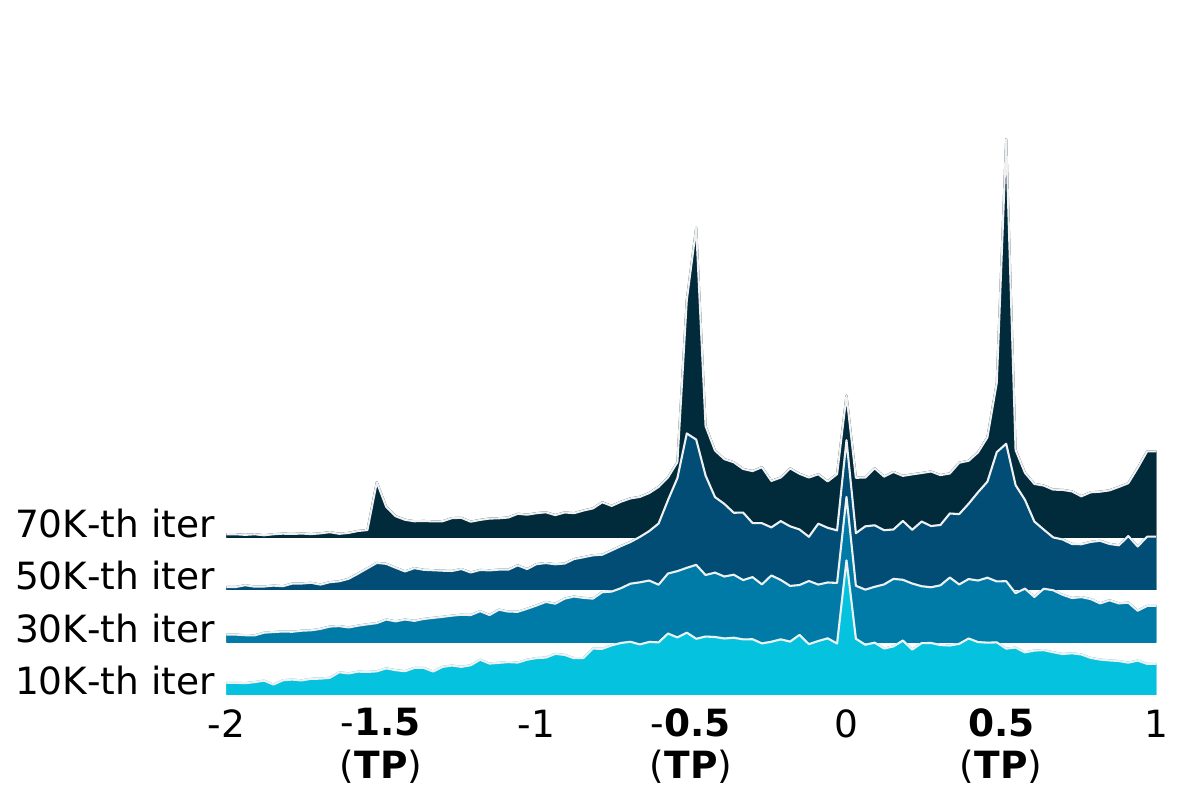}}
         \caption{Distributions of normalized latent weights.}
         \label{fig:appendix_discussion_distribution}
      \end{subfigure}
      \begin{subfigure}[t]{0.49\columnwidth}
         \centering
         \includegraphics[width=0.99\columnwidth]{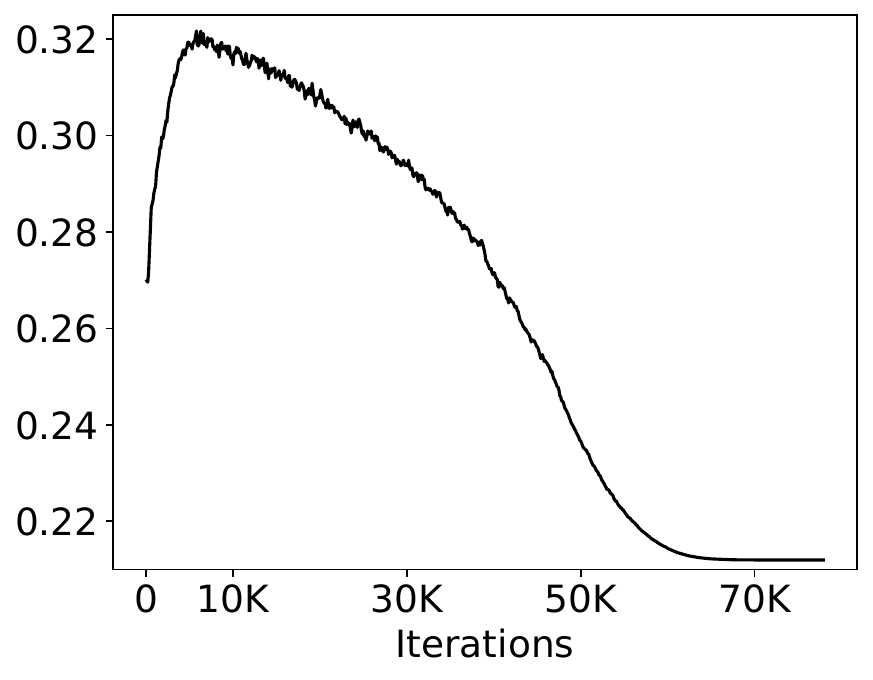}
         \caption{Average distances to the nearest transition points.}
         \label{fig:appendix_discussion_MD2TP}
      \end{subfigure}
   \end{center}
         \vspace{-0.2cm}
         \caption{Analysis on TR scheduling. We train ResNet-20~\cite{he2016deep} on CIFAR-100~\cite{krizhevsky2009learning} using SGDT, where we quantize both weights and activations with 2-bit representations. We visualize distributions of normalized latent weights in the 16$^\text{th}$ layer in~{\subref{fig:appendix_discussion_distribution}}, and average distances between normalized latent weights and the nearest transition points in~{\subref{fig:appendix_discussion_MD2TP}}. The transition points in~{\subref{fig:appendix_discussion_distribution}} are denoted by TPs in the x-axis. The top-1 test accuracy and average effective step sizes of quantized weights are shown by the red curves in Figs.~1d and~1c of the main paper, respectively. (Best viewed in color.)} 
   \label{fig:appendix_discussion}
 \end{figure*}

\section{Analysis} \label{sec:appendix_analysis}
\subsection{Analysis on TR scheduling}
We show in Fig.~\ref{fig:appendix_discussion} an in-depth analysis on how a TR scheduler works during QAT. We can see from Fig.~\ref{fig:appendix_discussion_tr} that the running TR~$K^t$ roughly follows the target TR~$R^t$, indicating that we can control the average effective step size of quantized weights~(the red curve in Fig.~1c of the main paper) by scheduling the target TR. This is possible because the TALR~$U^t$ is adjusted adaptively to match the running TR~$K^t$ with the target one~$R^t$~(Fig.~\ref{fig:appendix_discussion_talr}). We can see that the TALR~$U^t$ increases initially, since the running TR~$K^t$ is much smaller than the target TR~$R^t$. The TALR~$U^t$ then decreases gradually to reduce the number of transitions, following the target TR~$R^t$. Note that the TALR~$U^t$ approaches zero rapidly near the 50K-th iteration. To figure out the reason, we show in Figs.~\ref{fig:appendix_discussion_distribution} and~\ref{fig:appendix_discussion_MD2TP} distributions of normalized latent weights and their average distances to the nearest transition points, respectively. We can observe in Fig.~\ref{fig:appendix_discussion_distribution} that latent weights tend to be concentrated near the transition points of a quantizer as the training progresses, similar to the case in Sec.~4.1 in the main paper using a user-defined LR. This implies that transitions occur more frequently in later training iterations if we do not properly reduce the degree of parameter change for latent weights. In particular, we can see in Fig.~\ref{fig:appendix_discussion_MD2TP} the average distances between the normalized latent weights and the nearest transition points are relatively small after the 50K-th iteration. Under such circumstance, the TALR should become much smaller in order to reduce the running TR, as in the sharp decline around the 50K-th iteration. We can thus conclude that our approach adjusts the TALR by considering the distribution of the latent weights implicitly.

\begin{table*}[t]
   \setlength{\tabcolsep}{0.3em}
   \centering
   \small
   \caption{Quantitative comparison of optimization methods for QAT using a step decay scheduler on CIFAR-100/10~\cite{krizhevsky2009learning}. Both LRs and target TRs for SGD and SGDT are divided by 5 after every 50 and 100 epochs for ReActNet-18~\cite{liu2020reactnet} and ResNet-20~\cite{he2016deep}, respectively. Numbers in parentheses indicate accuracy drops compared to the models in Table~2 of the main paper trained with a cosine scheduler.}
     \begin{tabular}{C{1.6cm} C{2.7cm}  C{1.8cm}  C{1.8cm} C{2.7cm}  C{1.8cm} C{1.8cm}}
       \midrule
       \multirow{4}{*}[-3pt]{Optimizer} &  \multicolumn{3}{c}{CIFAR-100}                                              & \multicolumn{3}{c}{CIFAR-10}  \\
       \cmidrule(lr){2-4} \cmidrule(lr){5-7}
       \multicolumn{1}{c}{}             & \multicolumn{1}{c}{ReActNet-18}   & \multicolumn{2}{c}{ResNet-20}           & \multicolumn{1}{c}{ReActNet-18}   & \multicolumn{2}{c}{ResNet-20}  \\
       \multicolumn{1}{c}{}             & \multicolumn{1}{c}{(W32A1: 69.6)} & \multicolumn{2}{c}{(FP: 65.1)}          & \multicolumn{1}{c}{(W32A1: 91.3)} & \multicolumn{2}{c}{(FP: 91.1)} \\
                                        & 1/1                               & 1/1       & 2/2                         & 1/1                               & 1/1                & 2/2 \\
       \cmidrule(lr){1-1} \cmidrule(lr){2-2} \cmidrule(lr){3-4} \cmidrule(lr){5-5} \cmidrule(lr){6-7}
       SGD                              & 69.0 ($-$0.7)                     & 45.8 ($-$9.1)      & 61.3 ($-$2.8)      & 90.9 {\bf{($-$0.0)}}              & 82.9 ($-$2.3)      & 89.8 {\bf{($-$0.4)}} \\
       SGDT                             & \bf{71.9} ($-$0.3)                & \bf{55.3} ($-$0.5) & \bf{64.9} ($-$0.6) & \bf{93.0} ($-$0.0)                & \bf{85.2} ($-$0.4) & \bf{90.3} ($-$0.4) \\
       \cmidrule(lr){1-1} \cmidrule(lr){2-2} \cmidrule(lr){3-4} \cmidrule(lr){5-5} \cmidrule(lr){6-7}
       Adam                             & 68.1 ($-$1.4)                     & 49.9 ($-$4.9)      & 60.9  ($-$2.4)     & 82.5 ($-$7.9)                    & 82.4 ($-$2.4)      & 89.6 {\bf{($-$0.6)}} \\
       AdamT                            & \bf{71.7} ($-$0.1)                & \bf{54.4} ($-$1.5) & \bf{64.6} ($-$0.6) & \bf{92.9} ($-$0.0)                & \bf{85.3} ($-$0.4) & {\bf{90.3}} ($-$0.8) \\
       \midrule
     \end{tabular} \label{tab:appendix_cifar_step}
\end{table*}

\subsection{Comparison using a step decay scheduler}
We provide in Table~\ref{tab:appendix_cifar_step} a quantitative comparison between plain optimization methods~(SGD and Adam~\cite{kingma2014adam}) and ours~(SGDT and AdamT) using a step scheduler. We use the same training setting in Table~2 of the main paper, while replacing a cosine scheduler~\cite{loshchilov2016sgdr} with the step scheduler. Both LRs and target TRs are divided by 5 after every 50 and 100 epochs for ReActNet-18~\cite{liu2020reactnet} and ResNet-20~\cite{he2016deep}, respectively. We can see from this table that the plain optimizers using the step scheduler suffer from significant accuracy drops, compared to the cases for the the cosine scheduler. The main reason is that it is difficult to control the average effective step size of quantized weights using a user-defined LR, especially when the LR is decayed by the step scheduler. On the other hand, the optimizers using our TR scheduling technique are relatively robust to the step scheduler, showing less performance drops than the plain ones for most cases. We can also observe that the performance degrades more severely for the models using the ResNet-20~\cite{he2016deep} architecture, where the number of parameters is much less than that of the ReActNet-18~\cite{liu2020reactnet} architecture~(276K for ResNet-20 vs.~11,235K for ReActNet-18). This indicates that training a small quantized model with the step scheduler is more challenging, possibly because the scheduler can cause large parameter changes even at the end of the training.

\begin{figure}[t]
   \captionsetup[subfigure]{justification=centering}
   \begin{center}
      \begin{subfigure}[t]{0.47\columnwidth}
       \centering
       \includegraphics[width=1\columnwidth]{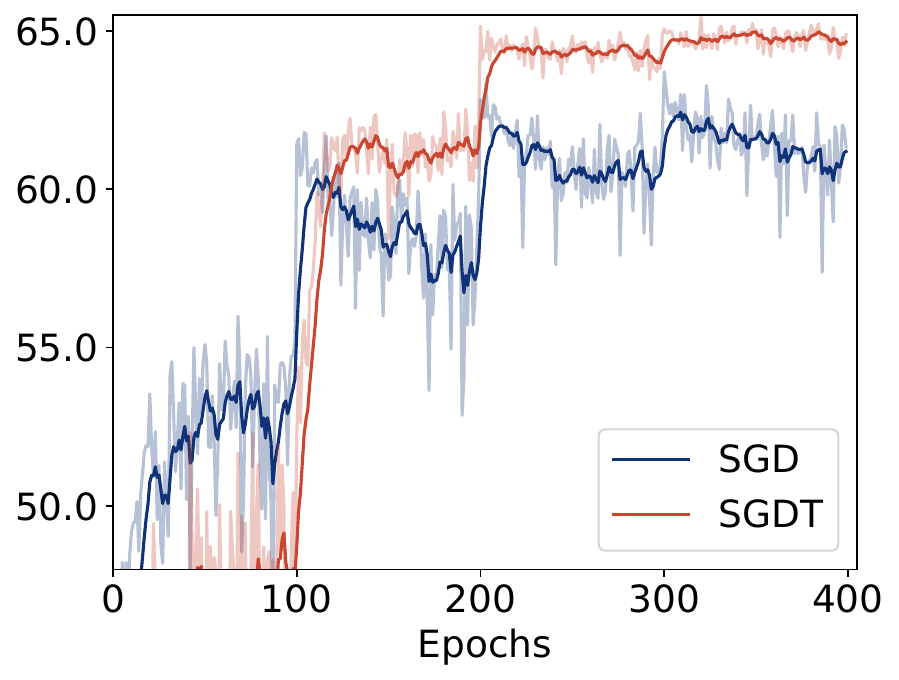}
       \caption{Test accuracy (\%).}
       \label{fig:appendix_scheduler_acc}
    \end{subfigure}
    \begin{subfigure}[t]{0.508\columnwidth}
       \centering
       \includegraphics[width=1\columnwidth]{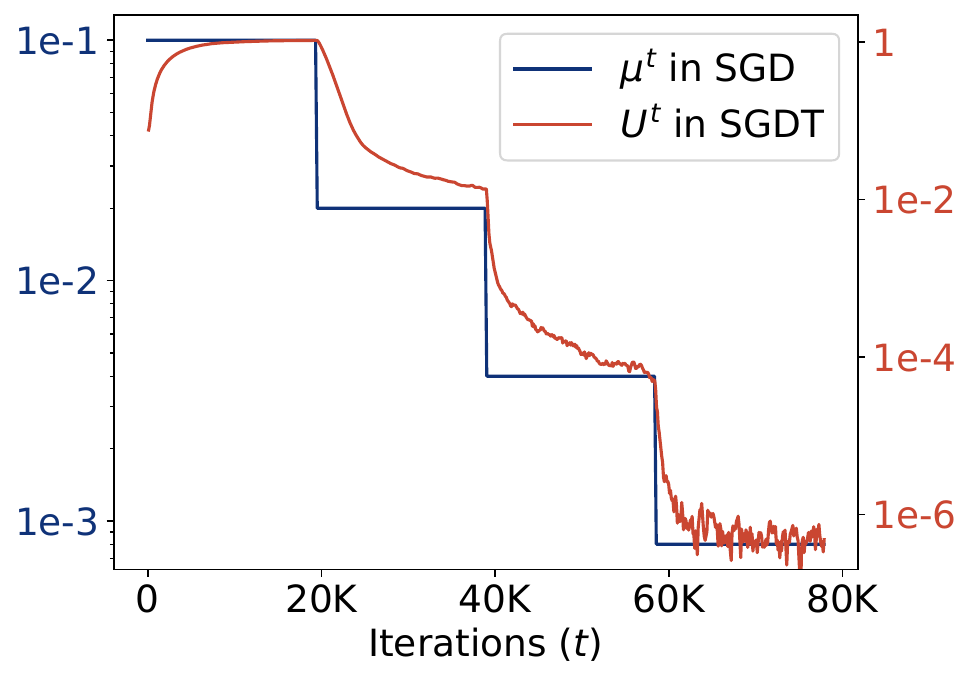}
       \caption{LR~$\mu^t$ and TALR~$U^t$.}
       \label{fig:appendix_scheduler_lr_talr}
    \end{subfigure}
   \end{center}
         \vspace{-0.3cm}
         \caption{Comparison between SGD and SGDT using a step scheduler. We train ResNet-20~\cite{he2016deep} with 2-bit weights and activations on CIFAR-100~\cite{krizhevsky2009learning}. Both the LR for SGD and target TR for SGDT are divided by 5 after every 100 epochs. For comparison, we show in~{\subref{fig:appendix_scheduler_lr_talr}} the LR~$\mu^t$ in SGD and TALR~$U^t$ in SGDT, where both are used for updating latent weights in Eqs.~(4) and~(12) of the main paper. (Best viewed in color.)}
   \label{fig:appendix_scheduler}
 \end{figure}

To further analyze the difference between the LR and TR scheduling techniques with the step scheduler, we compare in Fig.~\ref{fig:appendix_scheduler} the LR in SGD and TALR in SGDT during training, with corresponding test accuracies. We can observe from Fig.~\ref{fig:appendix_scheduler_acc} that SGDT provides better results in terms of both convergence rate and accuracy. As discussed in Sec.~4.1 of the main paper, latent weights approach transition points progressively during QAT, and thus it is difficult to adjust the number of transitions explicitly using a user-defined LR. This could be particularly problematic when the LR is fixed for a number of iterations. Recent QAT methods~\cite{liu2020reactnet,esser2019learned,lee2021network,yamamoto2021learnable,kim2021distance} circumvent this issue by decaying a LR to zero gradually,~\eg,~using the cosine annealing technique~\cite{loshchilov2016sgdr}, but this does not fully address the problem. On the other hand, our method is relatively robust to the use of the step scheduler. Since we optimize latent weights using a TALR, we can control the degree of parameter changes for quantized weights by scheduling the target TR in our method. In contrast to the LR, the TALR decreases gradually even when the scheduler does not alter the target TR~(Fig.~\ref{fig:appendix_scheduler_lr_talr}), confirming once more the fact that the TALR is adjusted adaptively considering the distribution of latent weights. This result demonstrates the robustness and generalization ability of our approach on various types of schedulers.

\begin{table}[t]
   \setlength{\tabcolsep}{0.285em}
      \centering
      \small
      \caption{Quantitative comparison of different TR factors~$\lambda$~(\ie,~initial target TRs). We quantize ResNet-20~\cite{he2016deep} using SGDT with 2-bit weights/activations, and report a top-1 test accuracy on CIFAR-100~\cite{krizhevsky2009learning}.}
      \adjustbox{max width = \columnwidth}{
         \begin{tabular}{C{1.8cm} c c c c c c c c c c}
            \midrule
            TR factor~$\lambda$      & 1e-3 & 2e-3 & 3e-3 & 4e-3 & 5e-3 & 6e-3 & 7e-3 & 8e-3 & 9e-3 & 1e-2 \\ \midrule 
            Test accuracy            & 62.5 & 64.2 & 64.3 & 65.3 & 65.5 & 65.1 & 63.1 & 63.6 & 63.6 & 64.0 \\
            \midrule
         \end{tabular}} \label{tab:appendix_different_initial_TRs}
         \vspace{0.15cm}
 \end{table}

\subsection{Analysis on initial target TRs}
To better understand how an initial target TR influences the quantization performance, we present in Table~\ref{tab:appendix_different_initial_TRs} a quantitative comparison of different TR factors~$\lambda$ adjusting the initial TRs. We train ResNet-20~\cite{he2016deep} models on CIFAR-100~\cite{krizhevsky2009learning} using 2-bit weights and activations with SGDT and report top-1 test accuracies. We can observe that the TR factors within a wide range of intervals~(\ie,~$[$4e-3, 6e-3$]$) provide satisfactory performance, outperforming the LR-based optimization strategy~(\ie, SGD in Table~2 of the main paper) by significant margins~(1.0$\sim$1.4). Similar to the LR, extremely large or small TR factors degrade the quantization performance, which could make the training unstable. Therefore, setting an appropriate value for the TR factor is crucial in the TR scheduling technique, analogous to determining an initial LR to train full-precision models.

\begin{table}[t]
   \setlength{\tabcolsep}{0.285em}
   \centering
   \small
   \caption{Quantitative comparison for plain optimization methods and our TR scheduling technique with various optimizers. We report a top-1 test accuracy on CIFAR-100~\cite{krizhevsky2009learning} using ResNet-20~\cite{he2016deep} with 2-bit weights/activations.}
   \adjustbox{max width = \columnwidth}{
   \begin{tabular}{C{0.72cm} c c c c c c c} 
      \midrule
               & SGD  & Adam & NAdam & Adamax & AdamW & RMSProp & Adagrad \\
         \midrule 
         Plain & 64.1 & 63.3 & 63.3  & 62.4   & 63.8  & 64.6    & 54.2 \\
         Ours  & 65.5 & 65.2 & 65.1  & 64.7   & 66.2  & 65.1    & 61.1 \\
      \midrule
   \end{tabular}
   }
   \label{tab:appendix_diverse_optims}
 \end{table}

\begin{table}[t]
   \setlength{\tabcolsep}{0.5em}
      \centering
      \footnotesize
      
      \caption{Quantitative comparison of quantized models trained with SGDT using different final target TRs. We report mean and standard deviation of top-1 test accuracies on CIFAR-100~\cite{krizhevsky2009learning} over five random runs, obtained with ResNet-20~\cite{he2016deep} using 2-bit weights and activations.}
      \vspace{-0.2cm}
      \adjustbox{max width = \columnwidth}{
       \begin{tabular}{c C{1.2cm} C{1.2cm} C{1.2cm} C{1.2cm}}
           \midrule
           Final target TR & 0 & 1e-5 & 1e-4 & 1e-3 \\ \midrule
           \multirow{2}{*}{\shortstack{Final target average\\effective step size}} & \multirow{2}{*}{0} & \multirow{2}{*}{5e-6} & \multirow{2}{*}{5e-5} & \multirow{2}{*}{5e-4} \\ 
           &&&&\\ \midrule 
           \multirow{2}{*}{Test accuracy} & 65.61 & 65.34  & 64.63 & 62.12 \\ 
                                          & ($\pm$0.21) & ($\pm$0.26) & ($\pm$0.59) & ($\pm$0.70) \\
           \midrule
       \end{tabular}
       }
        \label{tab:appendix_different_final_TRs}

 \end{table}

 \begin{table}[t]
   \centering
   \footnotesize
   \vspace{-0.4cm}
   \caption{Training time comparison on ImageNet with 4 A5000 GPUs, in terms of GPU hours. We quantize ResNet-18 except for AdamW/-T, where we quantize DeiT-T.}
   \label{tab:appendix_train_time}
   
%
   \begin{tabular}{C{0.72cm} c c c} 
      \midrule
               & SGD  & Adam & AdamW \\
         \midrule 
         Plain & 174 & 175 & 477 \\
         Ours  & 176 & 178 & 492  \\
      \midrule
   \end{tabular}
   
   \end{table}

\begin{figure*}[t]
   \captionsetup[subfigure]{justification=centering}
   \begin{center}
      \begin{subfigure}[t]{0.48\columnwidth}
         \centering
         \includegraphics[width=1\columnwidth]{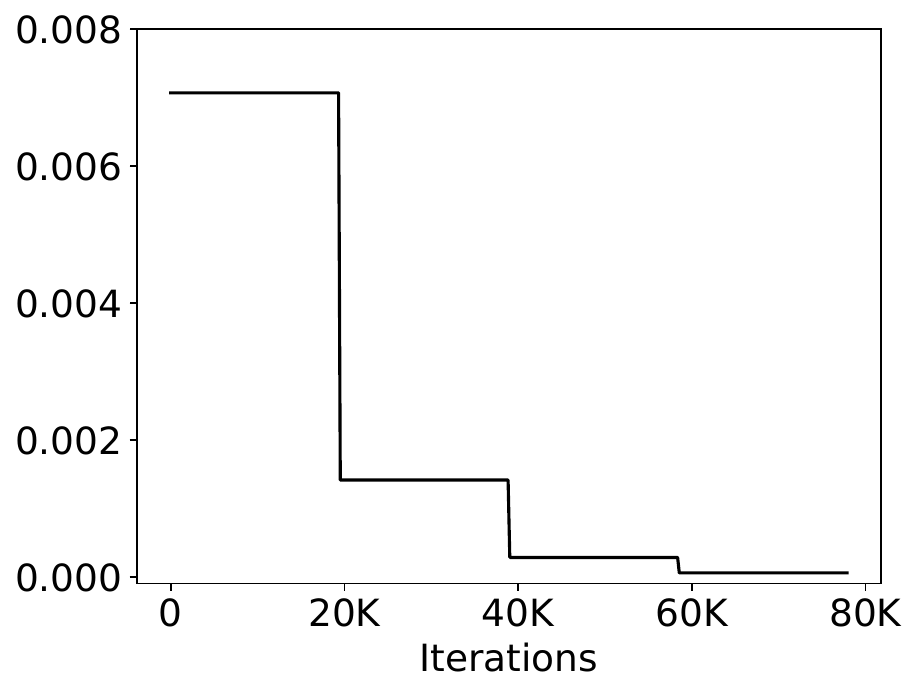}
         \caption{Target TR~$R^t$.} \label{fig:appendix_variants_target_TR}
      \end{subfigure}
      \begin{subfigure}[t]{0.48\columnwidth}
         \centering
         \includegraphics[width=1\columnwidth]{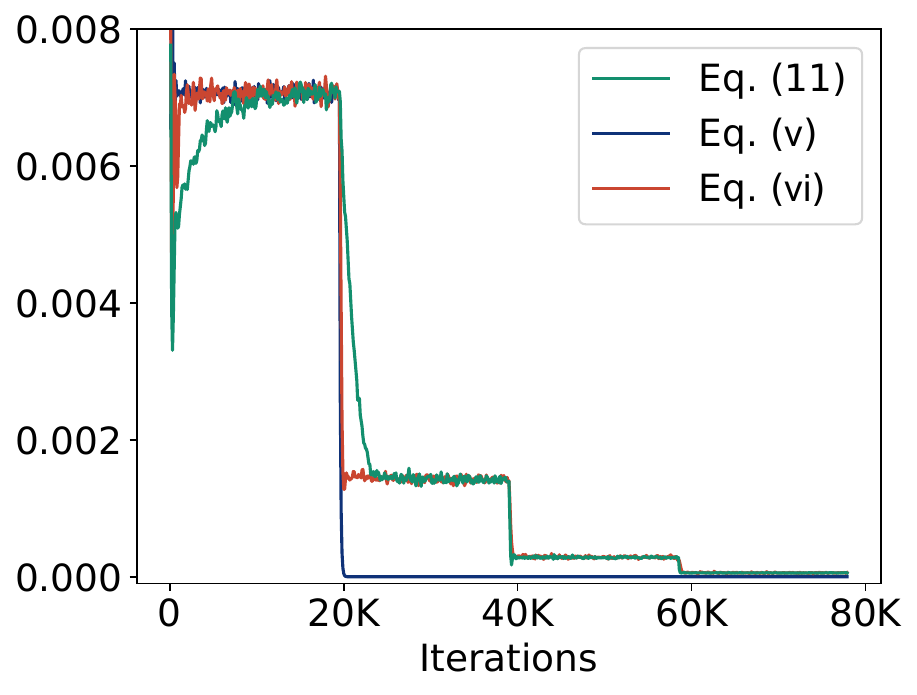}
         \caption{Running TR~$K^t$.} \label{fig:appendix_variants_running_TR}
      \end{subfigure}
      \begin{subfigure}[t]{0.45\columnwidth}
         \centering
         \raisebox{0.03cm}{\includegraphics[width=1\columnwidth]{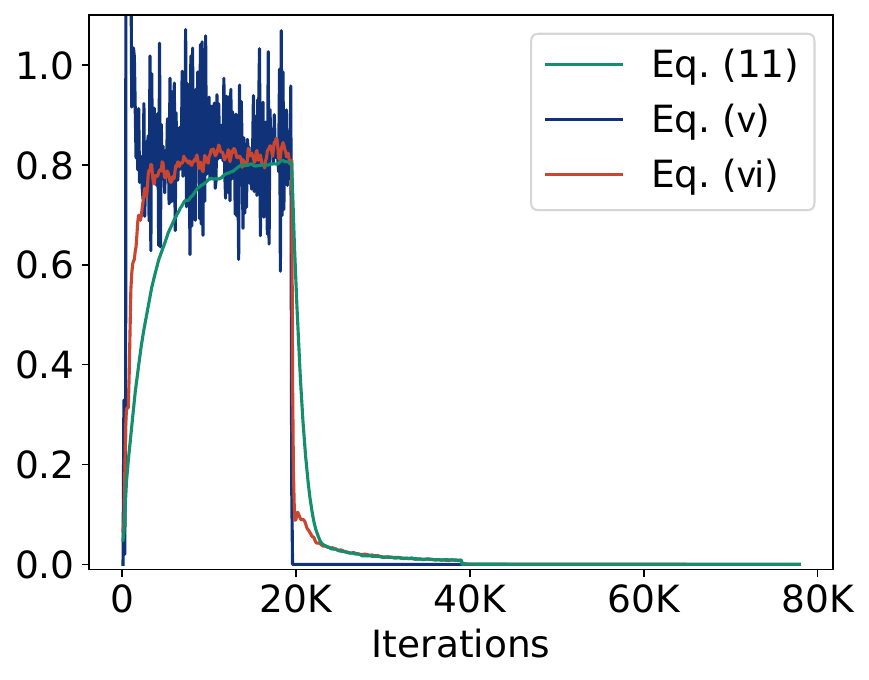}}
         \caption{TALR~$U^t$.} \label{fig:appendix_variants_TALR}
      \end{subfigure}
      \begin{subfigure}[t]{0.46\columnwidth}
         \centering
         \includegraphics[width=1\columnwidth]{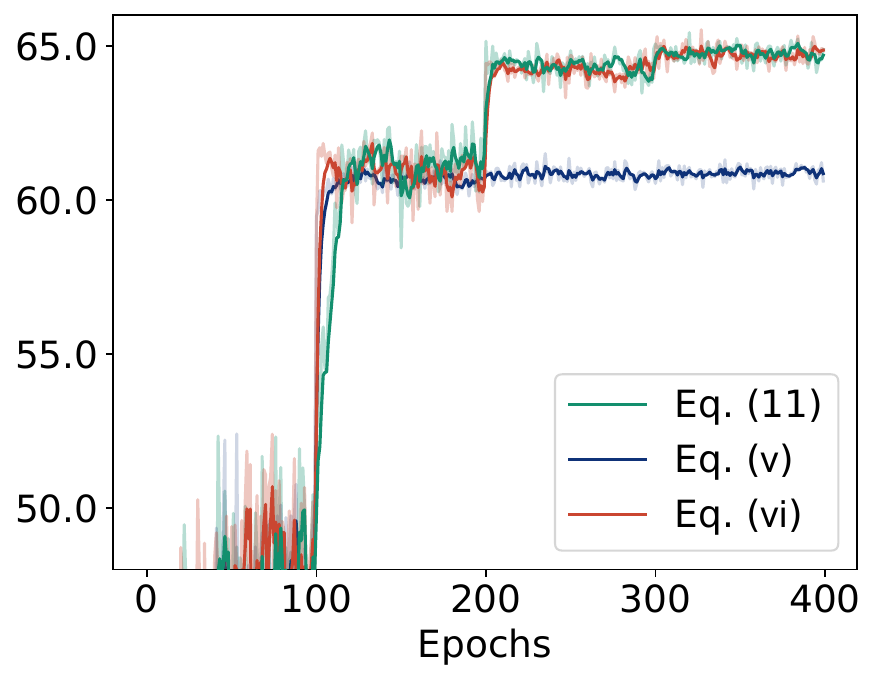}
         \caption{Test accuracy.} \label{fig:appendix_variants_acc}
      \end{subfigure}
   \end{center}
         \vspace{-0.2cm}
         \caption{Comparison of different update schemes for a TALR. We train ResNet-20~\cite{he2016deep} on CIFAR-100~\cite{krizhevsky2009learning} with 2-bit weights and activations, while adjusting the TALR based on different methods in Eqs.~(11),~\eqref{eq:variant1} and~\eqref{eq:variant2}. (Best viewed in color.)}
   \label{fig:appendix_variants}
 \end{figure*}

 \subsection{Comparison with different optimizers}
To verify the generalization ability of our TR scheduling technique to other optimizers, we compare in Table~\ref{tab:appendix_diverse_optims} the quantization performance between plain optimization methods and ours with various optimizers, including SGD, Adam~\cite{kingma2014adam}, NAdam~\cite{dozat2016nadam}, Adamax~\cite{kingma2014adam}, AdamW~\cite{loshchilov2017decoupled}, RMSProp~\cite{Tieleman2012}, and Adagrad~\cite{duchi2011adaptive}. Specifically, we train ResNet-20 on CIFAR-100~\cite{krizhevsky2009learning} with 2-bit weights and activations for comparison. For the optimizers other than SGD, we exploit the same hyperparameters as the ones for Adam in Table~2 of the main paper, while we change the weight decay to 1e-2 for AdamW. Note that AdamW decouples the weight decay from a parameter update step, suggesting that it requires a different hyperparameter for the weight decay. For RMSProp, we use the momentum gradient with a momentum value of 0.9, which is analogous to SGD. From the table, we can clearly see that the TR scheduling technique improves the quantization performance consistently across all optimizers. Our method achieves 0.5\%$\sim$6.9\% gains over the plain optimization methods, which demonstrates the effectiveness of our method and its generalization ability to various optimizers. In particular, AdamW coupled with our TR scheduling technique shows the best performance, indicating that we could further improve the quantization performance by carefully selecting an optimizer and corresponding hyperparameters.

\subsection{Training time comparison}
We compare in Table~\ref{tab:appendix_train_time} training time of optimization methods. Our method takes training time, similar to plain optimization methods, demonstrating its efficiency. The marginal overheads of ours is due to the computation of the TR and the adjustment of the TALR, both of which involve lightweight operations such as element-wise comparisons and scalar updates, which are computationally inexpensive compared to the overall training process.

\section{Discussion} \label{sec:appendix_discussion}
\subsection{Importance of reducing average effective step size}
Conventional optimization methods for full-precision models typically decay a LR to reduce the degree of parameter changes (\ie, average effective step sizes). This prevents full-precision weights from overshooting from a local optimum, and improves the convergence of the model~\cite{huang2017snapshot,kleinberg2018alternative}. Similarly, when training a quantized model, small average effective step sizes for quantized weights are preferred in later training iterations. Large changes in the quantized weights at the end of training could disturb convergence and degrade the quantization performance~\cite{nagel2022overcoming,park2020profit}, as shown in Figs.~1, 2, and \ref{fig:appendix_training_curves}. To further support this, we perform an experiment on the influence of the average effective step size of quantized weights on the quantization performance at the end of training. To this end, we train ResNet-20~\cite{he2016deep} on CIFAR-100~\cite{krizhevsky2009learning} with 2-bit weights and activations using SGDT for different final target TRs. We report in Table~\ref{tab:appendix_different_final_TRs} the mean and standard deviation of top-1 test accuracies over five random runs for each model. We can clearly see that the smaller average effective step sizes at the end of training provide better performance with less deviation, suggesting that the large step sizes in later training iterations disturb convergence. These results demonstrate once more the importance of the TR scheduling technique, since it is difficult to control the average effective step size of quantized weights, especially in later training iterations, with conventional LR scheduling in QAT.

\subsection{Comparison of update schemes for TALR} \label{sec:appendix_discussion_TALR_update}
We adjust a TALR adaptively to match a current running TR with a target one. To this end, the update rule for the TALR should meet the following criteria: (1)~If the running TR~$K^t$ is smaller than the target one~$R^t$, the TALR~$U^t$ should increase to allow more latent weights to pass transition points. (2)~If the running TR~$K^t$ is larger than the target one~$R^t$, the TALR~$U^t$ should decrease to reduce the number of transitions. (3)~If the current running TR~$K^t$ matches the target one~$R^t$, the TALR~$U^t$ should remain unchanged. We currently adjust the TALR in an additive manner in Eq.~(11) of the main paper, similar to the weight update in gradient-based optimizers. Here, we consider two alternative variants. First, we update the TALR in a multiplicative manner as follows:
\begin{equation} \label{eq:variant1}
   U^t = U^{t-1} \frac{R^t}{K^t}.
\end{equation}
While the multiplication allows fast adaptation for the TALR, it might be sensitive to outliers from the running TR~$K^t$ or drastic changes in the target TR~$R^t$. To stabilize the update, we adjust the TALR in a similar way to Eq~\eqref{eq:variant1}, but with momentum, as follows:
\begin{equation} \label{eq:variant2}
   U^t = m^\prime U^{t-1} + \left( 1-m^\prime \right) U^{t-1} \frac{R^t}{K^t},
\end{equation}
where~$m^\prime$ is a momentum hyperparameter, which is set to the same value as~$m$ in Eq.~(10) of the main paper for simplicity. To compare the update schemes, we analyze in Fig.~\ref{fig:appendix_variants} the QAT process of ResNet-20~\cite{he2016deep} on CIFAR-100~\cite{krizhevsky2009learning}. We can see from Figs.~\ref{fig:appendix_variants_target_TR} and~\ref{fig:appendix_variants_running_TR} that the running TRs~$K^t$ closely follow a target one~$R^t$ with the update schemes of Eqs.~(11) and~\eqref{eq:variant2}. The scheme in Eq.~\eqref{eq:variant1} fails to follow the target TR, resulting in the worst performance~(Fig.~\ref{fig:appendix_variants_acc}). The multiplicative update in Eq.~\eqref{eq:variant1} enables fast adaptation, but it is unstable and particularly problematic when the target TR~$R^t$ changes abruptly by a step scheduler. The momentum update in Eq.~\eqref{eq:variant2}, on the other hand, offers a good compromise between stability and adaptation speed, achieving the quantization performance comparable to the update scheme of Eq.~(11) of the main paper. Although we adopt the scheme of Eq.~(11) in our TR scheduling technique for conciseness, Fig.~\ref{fig:appendix_variants} suggests that how to adjust the TALR is not unique, and it can be updated in various ways. 
\subsection{Average effective step size}
The TR scheduling technique adjusts a TR of quantized weights~(\ie, Eq.~(5) of the main paper) explicitly, which in turn controls the degree of parameter changes for the quantized weights. As a variant of our method, we could instead formulate~$k^t$ in Eq.~(5) the main paper with the average effective step size of the quantized weights as follows:
\begin{equation} \label{eq:variant_ESS}
   k^t = \frac{\sum^{N}_{i=1} \left\vert w^t_q (i) - w^{t-1}_q(i) \right\vert}{N}.
\end{equation}
In this way, we could control the average effective step size of the quantized weights directly during training, similar to the TR in our method. Note that the effects of our method and its variant are nearly the same, if the quantized weights transit between adjacent discrete levels as discussed in Eq.~(9) of the main paper. This is always true for binary weights, or usually happens especially when the parameter updates for latent weights are small. When adopting the variant in Eq.~\eqref{eq:variant_ESS}, however, it could be difficult to set a hyperparameter~(\ie, an initial target value for the average effective step size) due to different scales in quantized weights. For example, several quantization methods~\cite{rastegari2016xnor,esser2019learned} provide quantized weights whose min-max ranges are different depending on layers in a model. In this case, we should use different search ranges for individual layers, to set initial target values, since the scales of average effective step sizes could vary in different layers. It is however computationally demanding to find an optimal initial target value for each layer separately. Accordingly, we focus on scheduling the TR rather than the average effective step size, which is more easy to use and generalizable to various quantization schemes.

\begin{table*}[!ht]
   \setlength{\tabcolsep}{0.7em}
   \centering
   \def\arraystretch{0.5}
   \caption{Hyperparameter settings in our experiments.}
   \begin{adjustbox}{width=1\linewidth,center}
   \begin{tabular}{C{2.3cm} C{2.5cm} C{1.5cm} C{1.5cm} C{1.4cm} C{1.1cm} C{1.8cm} C{2cm} C{1.7cm} C{2.5cm}}
      \midrule
      Dataset & Architecture & Optimizer & Epoch & Batch size &  LR & Weight decay & Decay scheduler & TR factor $\lambda$ & TR momentum $m$ \\
      \midrule
      \multirow{4}{*}[-9pt]{CIFAR-10/100} & \multirow{4}{*}[-9pt]{\shortstack{ResNet-20\\(ReActNet-18)}} & 
         SGD   & \multirow{4}{*}[-9pt]{\shortstack{400\\(200)}} & \multirow{4}{*}[-9pt]{256} & \multirow{2}{*}[-3pt]{1e-1} & \multirow{4}{*}[-9pt]{1e-4} & \multirow{4}{*}[-9pt]{cosine} & -    & -    \\ \cmidrule{3-3} \cmidrule{9-10}
      && SGDT  &                                                &                            &                             &                             &                               & 5e-3 & 0.99 \\ \cmidrule{3-3} \cmidrule{6-6} \cmidrule{9-10}
      && Adam  &                                                &                            & \multirow{2}{*}[-3pt]{1e-3} &                             &                               & -    & -   \\ \cmidrule{3-3} \cmidrule{9-10}
      && AdamT &                                                &                            &                             &                             &                               & 5e-3 & 0.99\\ 
      \midrule

      \multirow{12}{*}[-45pt]{ImageNet} & \multirow{4}{*}[-9pt]{\shortstack{ResNet-18}} & 
         SGD   & \multirow{4}{*}[-9pt]{120} & \multirow{4}{*}[-9pt]{256} & \multirow{2}{*}[-3pt]{1e-2} & \multirow{4}{*}[-9pt]{1e-4} & \multirow{4}{*}[-9pt]{cosine} & -    & -    \\ \cmidrule{3-3} \cmidrule{9-10}
      && SGDT  &                            &                            &                             &                             &                               & 1e-3 & 0.99 \\ \cmidrule{3-3} \cmidrule{6-6} \cmidrule{9-10}
      && Adam  &                            &                            & \multirow{2}{*}[-3pt]{1e-4} &                             &                               & -    & -    \\ \cmidrule{3-3} \cmidrule{9-10}
      && AdamT &                            &                            &                             &                             &                               & 1e-3 & 0.99 \\ \cmidrule{2-10}
      & \multirow{4}{*}[-9pt]{\shortstack{ReActNet-18}} & 
         SGD   & \multirow{4}{*}[-9pt]{120} & \multirow{4}{*}[-9pt]{256} & \multirow{2}{*}[-3pt]{1e-1} & \multirow{4}{*}[-9pt]{0}    & \multirow{4}{*}[-9pt]{linear} & -    & -    \\ \cmidrule{3-3} \cmidrule{9-10}
      && SGDT  &                            &                            &                             &                             &                               & 5e-4 & 0.99 \\ \cmidrule{3-3} \cmidrule{6-6} \cmidrule{9-10}
      && Adam  &                            &                            & \multirow{2}{*}[-3pt]{1e-3} &                             &                               & -    & -    \\ \cmidrule{3-3} \cmidrule{9-10}
      && AdamT &                            &                            &                             &                             &                               & 5e-4 & 0.99 \\ \cmidrule{2-10}
      & \multirow{4}{*}[-9pt]{\shortstack{MobileNetV2}} &
         SGD   & \multirow{4}{*}[-9pt]{120} & \multirow{4}{*}[-9pt]{256} & \multirow{2}{*}[-3pt]{1e-2} & \multirow{4}{*}[-9pt]{2.5e-5} & \multirow{4}{*}[-9pt]{cosine} & -    & -    \\ \cmidrule{3-3} \cmidrule{9-10}
      && SGDT  &                            &                            &                             &                               &                               & 1e-3 & 0.99 \\ \cmidrule{3-3} \cmidrule{6-6} \cmidrule{9-10}
      && Adam  &                            &                            & \multirow{2}{*}[-3pt]{1e-4} &                               &                               & -    & -    \\ \cmidrule{3-3} \cmidrule{9-10}
      && AdamT &                            &                            &                             &                               &                               & 1e-3 & 0.99 \\
      \cmidrule{2-10}
      & \multirow{2}{*}[-3pt]{DeiT-T/S} & 
         AdamW   & \multirow{2}{*}[-3pt]{300} & \multirow{2}{*}[-3pt]{256} & \multirow{2}{*}[-3pt]{3e-4} & \multirow{2}{*}[-3pt]{5e-2} & \multirow{2}{*}[-3pt]{cosine} & -    & -    \\ \cmidrule{3-3} \cmidrule{9-10}
      && AdamWT  &                                    &                           &                             &                             &                               & 1e-3 & 0.99 \\
      \midrule

      \multirow{2}{*}[-3pt]{MS COCO} & \multirow{2}{*}{\shortstack{RetinaNet w/\\ResNet backbone}} & 
         SGD   & \multirow{2}{*}[-3pt]{90K (iter.)} & \multirow{2}{*}[-3pt]{16} & \multirow{2}{*}[-3pt]{1e-2} & \multirow{2}{*}[-3pt]{1e-4} & \multirow{2}{*}[-3pt]{cosine} & -    & -    \\ \cmidrule{3-3} \cmidrule{9-10}
      && SGDT  &                                    &                           &                             &                             &                               & 1e-3 & 0.99 \\
      \midrule
   \end{tabular}
\end{adjustbox}\label{tab:appendix_hyperparam}
\end{table*}

\section{Implementation Details} \label{sec:appendix_implementation}
\subsection{Experimental settings} \label{sec:appendix_settings}
Following the experimental protocol in~\cite{zhou2016dorefa,liu2020reactnet,jung2019learning,lee2021network,yamamoto2021learnable}, we initialize network weights from pretrained full-precision models for MobileNetV2, ResNets, DeiTs and RetinaNet. Similarly, the weights in ReActNet-18 are initialized using the pretrained activation-only binarized model. We do not quantize the first and last layers. We use either plain optimizers using a LR~(SGD, Adam, and AdamW), or their variants using our TR scheduling technique~(SGDT, AdamT, and AdamWT), whose gradient terms are the same as SGD, Adam, and AdamW, respectively. Note that we use a TR scheduler to the latent weights only~(\ie, full-precision weights coupled with quantizers). Other parameters that are not quantized~(\eg, the weights in the first and last layers) are updated with the plain optimizers using a LR. We adopt the cosine scheduler~\cite{loshchilov2016sgdr} for both the LR and the target TR, decaying them to zero gradually. For ReActNet-18 on ImageNet, we use the linear scheduler, following the training scheme in~\cite{liu2020reactnet}. We exploit STE~\cite{bengio2013estimating} to propagate gradients in discretization functions. To set an initial value of the learnable scale parameter~$s$ in Eq.~(13) of the main paper, we follow the initialization technique in~\cite{lee2021network}. For an initial TALR, we use the same value as an initial LR for full-precision parameters. We also set~$\eta$ in Eq.~(11) of the main paper to the same value as an initial TALR~(\ie, $\eta=U^0$), adjusting the TALR according to its initial value,~\eg, updating the TALR more finely when the initial value is small. Note that the number of transition points increases as bit-widths of weights get larger, suggesting that a larger target TR is desirable for larger bit-widths. Accordingly, motivated by~\cite{esser2019learned} that uses different gradient scales according to bit-widths, we set the initial target TR to $\lambda \sqrt{b_w}$, where $\lambda$ is a TR factor, a hyperparameter to set, and $b_w$ is the bit-width of weights.

We summarize in Table~\ref{tab:appendix_hyperparam} hyperparameter settings for our experiments. If available, we follow the previous works~\cite{lin2017focal,esser2019learned,liu2020reactnet,lee2021network,nagel2022overcoming} to set hyperparameters~(\eg, number of training epochs/iterations, batch size, LR, weight decay, and decay scheduler). Different from the original work of RetinaNet~\cite{lin2017focal}, we do not freeze the ResNet~\cite{he2016deep} backbones to optimize quantized weights via QAT, and adopt the cosine decay scheduler~\cite{loshchilov2016sgdr}. For the models exploiting our TR scheduling technique, we fix the TR momentum constant~$m$ to 0.99 for simplicity. We choose the TR factor~$\lambda$ in a set of \{5e-3, 1e-3, 5e-4, 1e-4\}. When we train a scale parameter~$s$ in Eq.~(13) of the main paper, we set the LR to the value ten times smaller than the one for weight parameters~(\ie, the LR in Table~\ref{tab:appendix_hyperparam}), similar to the works of~\cite{jung2019learning,lee2021network}, for simplicity, instead of using the heuristic gradient scaling in LSQ~\cite{esser2019learned}. Note that when we apply our TR scheduling technique, we do not train the scale parameters~$s$ for weight quantizers and fix the initial values, since training them makes it difficult to control a TR of quantized weights. We will discuss this in more detail in the following section.

\begin{figure*}[t]
   \captionsetup[subfigure]{justification=centering}
   \begin{center}
      \begin{subfigure}[t]{0.485\columnwidth}
         \centering
         \raisebox{0.02cm}{\includegraphics[width=1\columnwidth]{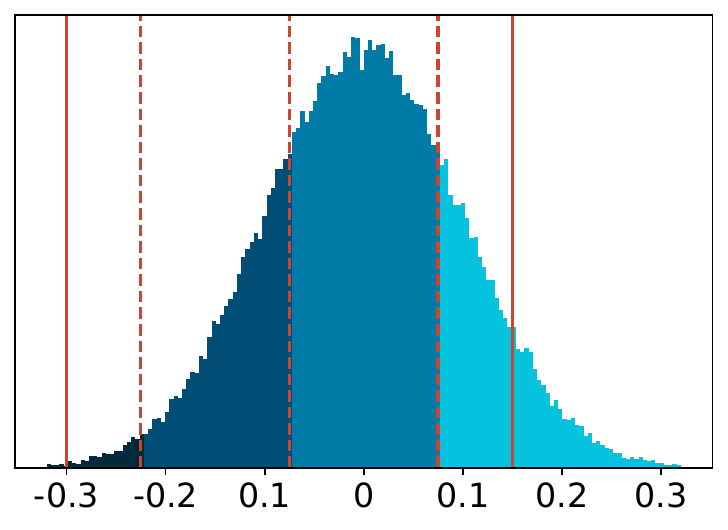}}
         \caption{Latent weights~${\bf{w}}$.} \label{fig:appendix_latent}
      \end{subfigure}
      \begin{subfigure}[t]{0.485\columnwidth}
         \centering
         \includegraphics[width=1\columnwidth]{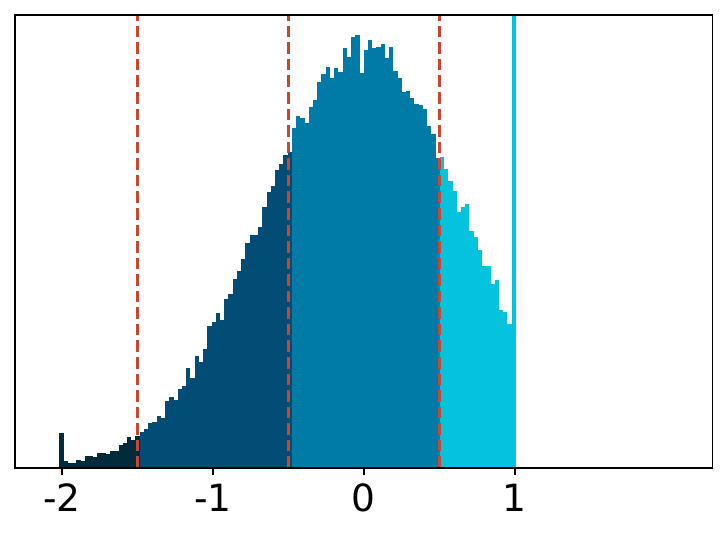}
         \caption{Normalized latent weights~${\bf{w}}_n$.} \label{fig:appendix_normalized}
      \end{subfigure}
      \begin{subfigure}[t]{0.485\columnwidth}
         \centering
         \includegraphics[width=1\columnwidth]{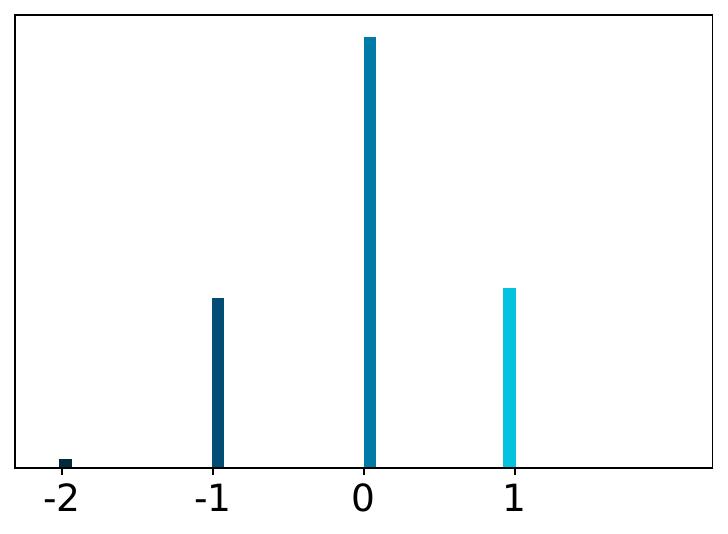}
         \caption{Discrete weights~${\bf{w}}_d$.} \label{fig:appendix_discrete}
      \end{subfigure}
      \begin{subfigure}[t]{0.485\columnwidth}
         \centering
         \includegraphics[width=1\columnwidth]{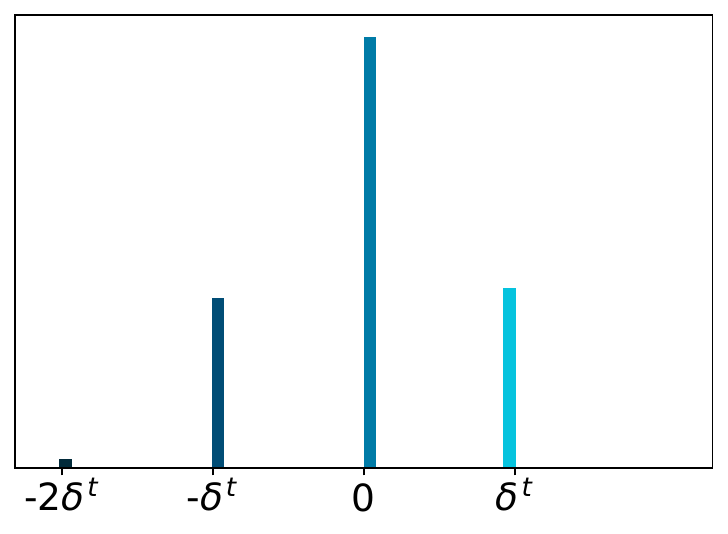}
         \caption{Quantized weights~${\bf{w}}_q$.} \label{fig:appendix_quantized}
      \end{subfigure}
   \end{center}
         \vspace{-0.3cm}
         \caption{Visualization of the quantization process for a 2-bit weight quantizer. Solid vertical lines in~\subref{fig:appendix_latent} indicate clipping points w.r.t latent weights, which adjust a quantization interval. Dashed vertical lines in~\subref{fig:appendix_latent} and~\subref{fig:appendix_normalized} represent transition points of the quantizer and a discretization function~(\ie, a round function) w.r.t the latent weights and normalized ones, respectively. The weights with the same color belong to the same discrete level of the quantizer. We denote by~$\delta^t$ in~\subref{fig:appendix_quantized} the distance between adjacent discrete levels of the quantizer at the~$t$-th iteration step. (Best viewed in color.)} 
   \label{fig:appendix_supple_quantizer}
 \end{figure*}

 \begin{figure}[t]
   \captionsetup[subfigure]{justification=centering}
   \begin{center}
      \begin{subfigure}[t]{0.485\columnwidth}
         \centering
         \includegraphics[width=1\columnwidth]{./figs/supple/latent_weights.pdf}
         \caption{Quantization interval w.r.t latent weights\\($s=0.3$).} \label{fig:appendix_s3}
      \end{subfigure}
      \begin{subfigure}[t]{0.485\columnwidth}
         \centering
         \includegraphics[width=1\columnwidth]{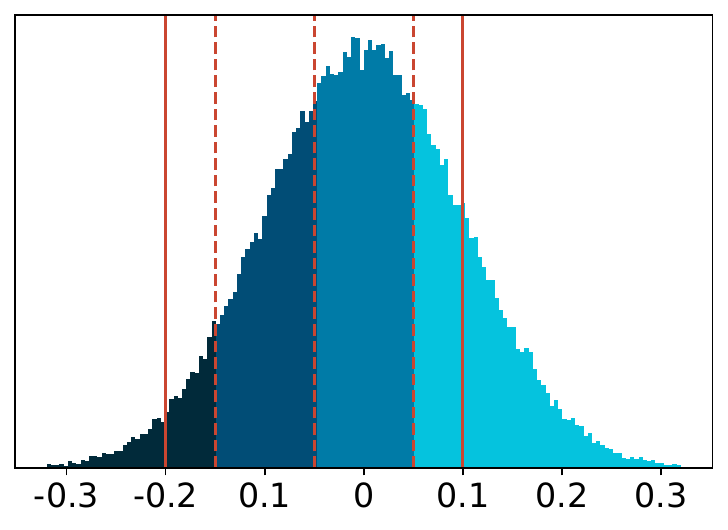}
         \caption{Quantization interval w.r.t latent weights\\($s=0.2$).} \label{fig:appendix_s2}
      \end{subfigure}
   \end{center}
         \vspace{-0.2cm}
         \caption{Comparison between quantization intervals for the same latent weights using different scale parameters. Solid and dashed vertical lines represent clipping and transition points of quantizers, respectively. The weights with the same color belong to the same discrete level of the quantizer. (Best viewed in color.)} \label{fig:appendix_supple_interval}
         \vspace{-0.4cm}
 \end{figure}

\section{Detailed Description of a Quantizer} \label{sec:appendix_supp_quantizer}
In this section, we explain the quantization process for a single quantizer. For simplicity, we consider a 2-bit weight quantizer based on Eq.~(13) of the main paper, and use latent weights randomly drawn from the zero-mean Gaussian distribution with a standard deviation of 0.1. In the following, we describe a detailed procedure of quantization according to the steps described in Eqs.~(1)-(3) of the main paper. We visualize in Fig.~\ref{fig:appendix_supple_quantizer} an overall quantization process for the quantizer. First, the latent weights~$\bf{w}$~(Fig.~\ref{fig:appendix_latent}) are normalized and clipped by a normalization function, producing normalized latent weights~${\bf{w}}_n$~(Fig.~\ref{fig:appendix_normalized}):
\begin{equation} \label{eq:supp_normalize}
  {\bf{w}}_n = \text{clip} \left( \frac{\gamma {\bf{w}}}{s}, \alpha, \beta \right),
\end{equation}
where we set~$\alpha=-2$, $\beta=1$, and $\gamma=2$ for 2-bit weight quantization. $s$ is a learnable scale parameter adjusting a quantization interval. In Fig.~\ref{fig:appendix_supple_quantizer}, we set~$s=0.3$ for the purpose of visualization. Second, the normalized latent weights~${\bf{w}}_n$ are converted to discrete ones~${\bf{w}}_d$~(Fig.~\ref{fig:appendix_discrete}) using a round function:
\begin{equation} \label{eq:supp_discrete}
  {\bf{w}}_d = \left\lceil  {\bf{w}}_n  \right\rfloor.
\end{equation}
Lastly, the discrete weights~${\bf{w}}_d$ are fed into a de-normalization function that applies post-scaling and outputs quantized weights~${\bf{w}}_q$~(Fig.~\ref{fig:appendix_quantized}):
\begin{equation} \label{eq:supp_post}
  {\bf{w}}_q = \frac{1}{\gamma}  {\bf{w}}_d.
\end{equation}
Note that the post-scaling in~Eq.~\eqref{eq:supp_post} is fixed, dividing the discrete weights with a constant value~$\gamma$ of 2. In this case, the distance between the adjacent discrete levels of the quantizer~(\ie,~$\delta^t$ in Fig.~\ref{fig:appendix_quantized}) is fixed for all training iterations, \ie,~$\delta^t = 0.5$ for all~$t$.

\subsection{Counting transitions}
In Eq.~(5) in the main paper, we count the number of transitions by observing whether discrete weights,~\ie,~integer numbers resulting from a round or a signum function~(\eg,~${\bf{w}}_d$ in Eq.~\eqref{eq:supp_discrete}), are changed or not after a single parameter update. As an example, suppose a case that a quantized weight at the $t$-th iteration step~$w_q^t$ belongs to the first level of the quantizer, \eg,~$w_q^t = -2\delta^t$ in Fig.~\ref{fig:appendix_quantized}, where corresponding discrete weight~$w_d^t$ in Fig.~\ref{fig:appendix_discrete} is $-2$. If the quantized weight transits from the first to the second level of the quantizer after a parameter update~(\ie, $w_q^{t+1}=-\delta^{t+1}$), we can detect the transition using the discrete weight, since it is changed from~$-2$ to~$-1$. Similarly, if the quantized weight remains in the same level after a parameter update~(\ie, $w_q^{t+1}=-2\delta^{t+1}$), we can say that the transition does not occur, because the discrete weight retains the same value. Note that we could use quantized weights~${\bf{w}}_q$ instead of discrete weights~${\bf{w}}_d$ for counting the number of transitions in Eq.~(5) in the main paper, only when~$\delta^t$ is fixed for all training iterations~(\eg,~as in our quantizer in Eq.~(13) of the main paper). Otherwise this could be problematic. For example, even if a quantized weight does not transit the discrete level after a parameter update, \eg, $w_q^{t} = -2\delta^t$ and $w_q^{t+1} = -2\delta^{t+1}$, the quantized weight can be changed if $\delta^t$ and $\delta^{t+1}$ are not the same. This indicates that we cannot detect a transition with the condition of $\mathbb{I}\left[ w^{t+1}_q \neq w^{t}_q \right]$, since the statement~($w^{t+1}_q \neq w^{t}_q$) could be always true, regardless of whether a transition occurs or not, if~$\delta^{t+1} \neq \delta^t$ for all training iterations. Consequently, we count the number of transitions using discrete weights in Eq.~(5) in the main paper, which is valid for general quantizers.

\subsection{Quantization interval}
Following the recent state-of-the-art quantization methods~\cite{jung2019learning,esser2019learned,lee2021network}, we introduce in Eq.~(13) of the main paper~(or in Eq.~\eqref{eq:supp_normalize}) a learnable scale parameter~$s$. Given that $\alpha$, $\beta$ and $\gamma$ in Eq.~(13) of the main paper are bit-specific constants, the scale parameter~$s$ is the only factor that controls a quantization interval~(\ie, a clipping range) w.r.t quantizer inputs. We can see from Fig.~\ref{fig:appendix_latent} that the scale parameter~($s=0.3$) is responsible for the positions of clipping points w.r.t latent weights~(solid vertical lines in Fig.~\ref{fig:appendix_latent}). It also decides transition points accordingly~(dashed vertical lines in Figs.~\ref{fig:appendix_latent}), since the points are set by splitting the clipping range uniformly. This suggests that different scale parameters would give different sets of transition points. To verify this, we compare in Fig.~\ref{fig:appendix_supple_interval} the quantization intervals using different scale parameters~$s$ for the same latent weights. We can see that the quantization interval shrinks if a smaller scale parameter~($s=0.2$) is used, and the transition points are altered consequently. This again suggests that transitions could occur if the scale parameter~$s$ is changed during training, even when the latent weights are not updated. For example, a latent weight of $-0.2$ in Fig.~\ref{fig:appendix_s3} belongs to the second level of the quantizer, whereas that in Fig.~\ref{fig:appendix_s2} belongs to the first level. Within our TR scheduling technique, we attempt to control the number of transitions by updating latent weights with a TALR, but the transitions could also be affected by the scale parameter. For this reason, we do not train the scale parameters in weight quantizers, when the TR scheduling technique is adopted, fixing the transition points of the quantizers and controlling the transitions solely with the latent weights.






\end{document}